\documentclass[runningheads]{llncs}
\usepackage{graphicx}

\usepackage{tikz}
\usepackage{comment}
\usepackage{multirow}
\usepackage{amsmath,amssymb} 
\usepackage{color}

\usepackage[accsupp]{axessibility}  

\usepackage{capt-of}
\usepackage{xspace}
\usepackage{booktabs}

\newcommand{\vol}{V}
\newcommand{\pose}{\phi_i}
\newcommand{\rot}{R_i}
\newcommand{\trans}{\mathbf{t}_{i}}
\newcommand{\projgt}{Y_i}
\newcommand{\enc}{\xi}
\newcommand{\psf}{P_i}
\newcommand{\ctf}{C_i}
\newcommand{\fvol}{\hat{V}}
\newcommand{\fprojgt}{\hat{Y}_i}

\newcommand{\fprojpred}{\hat{X}_i}
\newcommand{\surf}{Q_i}
\newcommand{\Fthree}{\mathcal{F}_{3\text{D}}}
\newcommand{\Ftwo}{\mathcal{F}_{2\text{D}}}
\newcommand{\slice}{\mathcal{S}_{i}}
\newcommand{\sidelen}{L}
\newcommand{\noise}{\eta_i}
\newcommand{\fnoise}{\hat{\eta}_i}

\newcommand{\batch}{\mathcal{B}}
\newcommand{\symloss}{\mathcal{L}_{\text{sym}}}
\newcommand{\simba}{\Gamma_{\xi,\theta}}
\newcommand{\flip}{\mathcal{R}_\pi}
\newcommand{\cryoSIMBA}{cryoAI\xspace}
\newcommand{\CryoSIMBA}{CryoAI\xspace}

\newcommand{\ra}[1]{\renewcommand{\arraystretch}{#1}}

\begin{document}
\pagestyle{headings}
\mainmatter
\def\ECCVSubNumber{3418}  

\title{\CryoSIMBA: Amortized Inference of Poses for Ab Initio Reconstruction of 3D Molecular Volumes from Real Cryo-EM Images}

\titlerunning{CryoAI}
%
\author{
Axel Levy$^{1,2,*}$\and
Fr\'{e}d\'{e}ric Poitevin$^{1,*}$\and
Julien Martel$^{2,*}$\and
Youssef Nashed$^{3}$\and
Ariana Peck$^{1}$\and
Nina Miolane$^{4}$\and
Daniel Ratner$^{3}$\and
Mike Dunne$^{1}$\and
Gordon Wetzstein$^{2}$
}
\authorrunning{A. Levy, F. Poitevin, J. Martel et al.}
%
\institute{
$^{1}$LCLS, SLAC National Accelerator Laboratory, Menlo Park, CA, USA\\
$^{2}$Stanford University, Department of Electrical Engineering, Stanford, CA, USA\\
$^{3}$ML Initiative, SLAC National Accelerator Laboratory, Menlo Park, CA, USA\\
$^{4}$University of California Santa Barbara, Department of Electrical and Computer Engineering, Santa Barbara, CA, USA
}
\maketitle

\begin{abstract}
Cryo-electron microscopy (cryo-EM) has become a tool of fundamental importance in structural biology, helping us understand the basic building blocks of life. The algorithmic challenge of cryo-EM is to jointly estimate the unknown 3D poses and the 3D electron scattering potential of a biomolecule from millions of extremely noisy 2D images. 
Existing reconstruction algorithms, however, cannot easily keep pace with the rapidly growing size of cryo-EM datasets due to their high computational and memory cost. We introduce \cryoSIMBA, an \emph{ab initio} reconstruction algorithm for homogeneous conformations that uses direct gradient-based optimization of particle poses and the electron scattering potential from single-particle cryo-EM data. \CryoSIMBA combines a learned encoder that predicts the poses of each particle image with a physics-based decoder to aggregate each particle image into an implicit representation of the scattering potential volume. This volume is stored in the Fourier domain for computational efficiency and leverages a modern coordinate network architecture for memory efficiency. Combined with a symmetric loss function, this framework achieves results of a quality on par with state-of-the-art cryo-EM solvers for both simulated and experimental data, one order of magnitude faster for large datasets and with significantly lower memory requirements than existing methods. 

\keywords{Cryo-electron Microscopy, Neural Scene Representation}
\end{abstract}

\section{Introduction}\label{sec:intro}

\begin{figure}[t!]
\centering
\includegraphics[width=\textwidth]{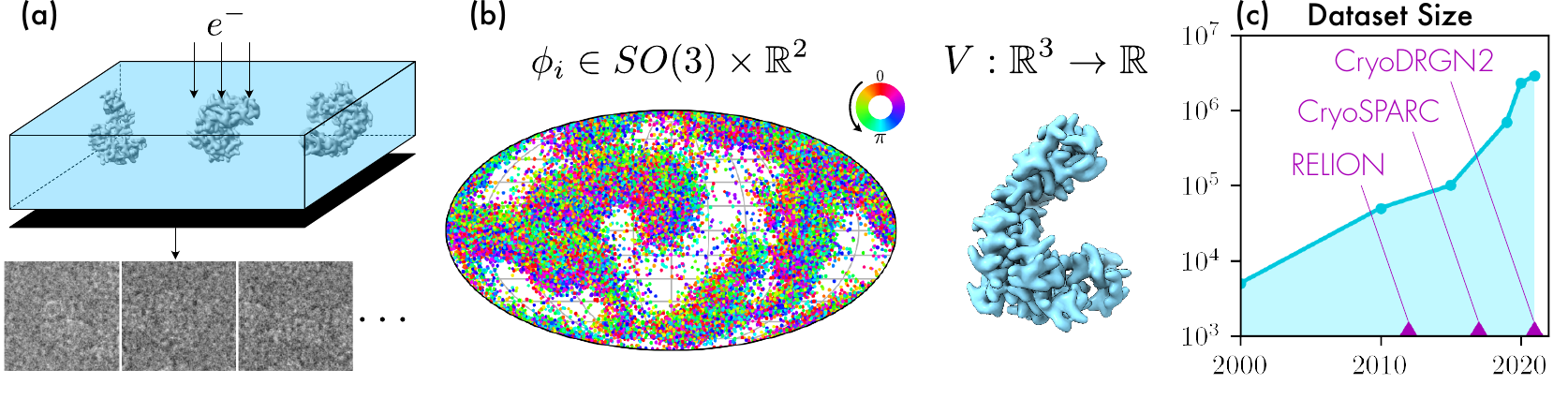}
\caption{(a) (Top) Illustration of a cryo-EM experiment. Molecules are frozen in a random orientation and their electron scattering potential (i.e., volume) $\vol$ interacts with an electron beam imaged on a detector. (Bottom) Noisy projections (i.e., particles) of $\vol$ selected from the full micrograph measured by the detector. (b) Output of a reconstruction algorithm: poses $\pose$ and volume $V$. Each pose is characterized by a rotation in $SO(3)$ (hue represents in-plane rotation) and a translation in $\mathbb{R}^2$ (not shown). An equipotential surface of $\vol$ is shown on the right. (c) Evolution of the maximum number of images collected in one day \cite{namba_2022} and established and emerging state-of-the-art reconstruction methods.} 
\label{fig:cryoembn}
\end{figure}

Understanding the 3D structure of proteins and their associated complexes is crucial for drug discovery, studying viruses, and understanding the function of the fundamental building blocks of life. Towards this goal, cryo-electron microscopy (cryo-EM) of isolated particles has been developed as the go-to method for imaging and studying molecular assemblies at near-atomic resolution~\cite{Kulbrandt:2014,Nogales:2016,Renaud:2018}. In a cryo-EM experiment, a purified solution of the molecule of interest is frozen in a thin layer of vitreous ice, exposed to an electron beam, and randomly oriented projections of the electron scattering potential (i.e., the volume) are imaged on a detector (Fig.~\ref{fig:cryoembn} (a)). These raw \emph{micrographs} are then processed by an algorithm that reconstructs the volume and estimates the unknown pose, including orientation and centering shift, of each particle extracted from the micrographs (Fig.~\ref{fig:cryoembn} (b)).

Recent advances in sample preparation, instrumentation, and data collection capabilities have resulted in very large amounts of data being recorded for each cryo-EM experiment \cite{baldwin_2018,namba_2022} (Fig.~\ref{fig:cryoembn}~(c)). Millions of noisy (images of) particles, each with an image size on the order of $100^2$--$400^2$ pixels, need to be processed by the reconstruction algorithm to jointly estimate the pose of each particle and the unknown volume. Most existing algorithms that have been successful with experimental cryo-EM data address this problem using a probabilistic approach that iteratively alternates between updating the volume and the estimated poses~\cite{scheres_relion_2012,punjani_cryosparc_2017,zhong_cryodrgn_2021,zhong_cryodrgn2_2021}. The latter ``orientation matching'' step, however, is computationally expensive, requiring an exhaustive search in a 5-dimensional space ($\pose\in SO(3)\times\mathbb{R}^2$) for each particle. In spite of using smart pose search strategies and optimization schedules, the orientation matching step is the primary bottleneck of existing cryo-EM reconstruction algorithms, requiring hours to estimate a single volume and scaling poorly with increasing dataset sizes.

We introduce \cryoSIMBA, a technique that uses direct gradient-based optimization to jointly estimate the poses and the electron scattering potential of a non-deformable molecule (\emph{homogeneous} reconstruction). Our method operates in an unsupervised manner over a set of images with an encoder--decoder pipeline. The encoder learns a discriminative model that associates each particle image with a pose and the decoder is a generative physics-based pipeline that uses the predicted pose and a description of the volume to predict an image. The volume is maintained by an implicit, i.e., neural network--parameterized, representation in the decoder, and the image formation model is simulated in Fourier space, thereby avoiding the approximation of integrals via the Fourier-slice theorem (see Sec.~\ref{subsec:imf}). 
By learning a mapping from images to poses, \cryoSIMBA avoids the computationally expensive step of orientation matching that limits existing cryo-EM reconstruction methods. Our approach thus amortizes over the size of the dataset and provides a scalable approach to working with modern, large-scale cryo-EM datasets. We demonstrate that \cryoSIMBA performs homogeneous reconstructions of a comparable resolution but with nearly one order of magnitude faster runtime than state-of-the-art methods using datasets containing millions of particles. 

Specifically, our contributions include
\begin{itemize}
    \item a framework that learns to map images to particle poses while reconstructing an electron scattering potential for homogeneous single-particle cryo-EM; 
    \item demonstration of reconstruction times and memory consumption that amortize over the size of the dataset, with nearly an order of magnitude improvement over existing algorithms on large datasets;
    \item formulations of a symmetric loss function and an implicit Fourier-domain volume representation that enable the high-quality reconstructions we show.
\end{itemize}

Source code is available at \url{https://github.com/compSPI/cryoAI}.

\section{Related Work}\label{sec:previous_work}

Estimating the 3D structure of an object from its 2D projections with known orientations is a classical problem in tomography and has been solved using backprojection-based methods \cite{hertle_problem_1981,rudin_nonlinear_1992} or compressive sensing--style solvers \cite{candes_robust_2006,donoho_compressed_2006}. In cryo-EM, the reconstruction problem is complicated by several facts: (1) the poses of the unknown object are also unknown for all projections; (2) the signal-to-noise ratio (SNR) is extremely low (around $-20~\text{dB}$ for experimental datasets~\cite{bepler2020topaz,bendory2020single}); (3) the molecules in a sample can deform and be frozen in various (unknown) conformations. Unlike homogeneous reconstruction methods, heterogeneous methods take into account the deformations of the molecule and reconstruct a discrete set or a low-dimensional manifold of conformations. Although they give more structural information, most recent heterogeneous methods \cite{zhong_cryodrgn_2021,punjani_3d_2021,zhong_exploring_2021,chen_deep_2021} assume the poses to be known. For each particle $i$, a pose $\pose$ is defined by a rotation $\rot\in SO(3)$ and a translation $\trans\in\mathbb{R}^2$. In this work, we do not assume the poses to be known and aim to estimate the electron scattering function $\vol$ of a unique underlying molecule in a homogeneous setting. We classify previous work on pose estimation into two inference categories \cite{donnat_deep_2022}: non-amortized and amortized.

\textbf{Non-amortized Inference} refers to a class of methods where the posterior distribution of the poses $p(\pose|\projgt, \vol)$ is computed independently for each image $\projgt$. Common-line approaches \cite{vainshtein_determination_1986,singer_detecting_2010,wang_orientation_2013,greenberg_common_2017,pragier_common_2019,zehni_joint_2020}, projection-matching strategies \cite{penczek_ribosome_1994,baker_model-based_1996} and Bayesian formulations \cite{mallick_structure_2006,dempster_maximum_1977,sigworth_maximum-likelihood_1998,punjani_cryosparc_2017} belong to this category. The software package RELION \cite{scheres_relion_2012} widely popularized the Bayesian approach by performing Maximum-A-Posteriori (MAP) optimization through Expectation--Maximization (EM). Posterior distributions over the poses (and the optional \emph{conformational} states) are computed for each image in the expectation step and all frequency components of the volume are updated in the maximization step, which makes the approach computationally costly. The competing software cryoSPARC \cite{punjani_cryosparc_2017} proposed to perform MAP optimization jointly using stochastic gradient descent (SGD) to optimize the volume $\vol$ and branch-and-bound algorithms~\cite{lawler1966branch} to estimate the poses $\pose$. While a gradient-based optimization scheme for $\vol$ circumvents the costly updates in the maximization step of RELION, a pose must be estimated for each image by aligning each 2D projection $\projgt$ with the estimated 3D volume $\vol$. Although branch-and-bound algorithms can accelerate the pose search, this step remains computationally expensive and is one of the bottlenecks of the method in terms of runtime. Ullrich \textit{et al.} \cite{ullrich_differentiable_2019} proposed a variational and differentiable formulation of the optimization problem in the Fourier domain. Although they demonstrated that their method can estimate the volume when poses are known, they also showed that jointly optimizing the pose posterior distributions by SGD fails due to the high non-convexity of the problem. Instead of parameterizing the volume with a 3D voxel array, Zhong \textit{et al.} proposed in cryoDRGN \cite{zhong_reconstructing_2019,zhong_cryodrgn_2021,zhong_cryodrgn2_2021} to use a coordinate-based representation (details in Sec.~\ref{subsec:FN}) to directly approximate the electron scattering function in Fourier space. Their neural representation takes 3D Fourier coordinates and a latent vector encoding the conformational state as input, therefore accounting for continuous deformations of the molecule. The latest published version of cryoDRGN~\cite{zhong_cryodrgn_2021} reports excellent results on the reconstruction of conformation heterogeneities but assumes the poses to be determined by a consensus reconstruction. Poses are jointly estimated with $\vol$ in cryoDRGN-BNB~\cite{zhong_reconstructing_2019} and cryoDRGN2~\cite{zhong_cryodrgn2_2021}, but in spite of a frequency-marching strategy, the use of a branch-and-bound algorithm and a later introduced multi-resolution approach the global 5D pose search remains the most computationally expensive step in their pipeline. 

\textbf{Amortized Inference} techniques, on the other hand, learn a parameterized function $q_\enc(\projgt)$ that approximates the posterior distribution of the poses $p(\pose|\projgt, \vol)$ \cite{gershman2014amortized}. At the expense of optimizing the parameter $\enc$, these techniques avoid the orientation matching step which is the main computational bottleneck in non-amortized methods. Lian \textit{et al} \cite{lian_end--end_2022} demonstrated the possibility of using a convolutional neural network to approximate the mapping between cryo-EM images and orientations, but their method cannot perform end-to-end volume reconstruction. In cryoVAEGAN \cite{miolane_estimation_2020}, Miolane \textit{et al.} showed that the in-plane rotation could be disentangled from the contrast transfer function (CTF) parameters in the latent space of an encoder. Rosenbaum \textit{et al.} \cite{rosenbaum_inferring_2021} were the first to demonstrate volume reconstruction from unknown poses in a framework of amortized inference. In their work, distributions of poses and conformational states are predicted by the encoder of a Variational Autoencoder (VAE) \cite{kingma2019introduction}. In their model-based decoder, the predicted conformation is used to deform a base backbone frame of Gaussian blobs and the predicted pose is used to make a projection of these blobs. The reconstructed image is compared to the measurement in order to optimize the parameters of both the encoder and the decoder. While this method is able to account for conformational heterogeneity in a dataset, it requires \textit{a priori} information about the backbone frame. CryoPoseNet \cite{nashed_cryoposenet_2021} proposed a non-variational autoencoder framework that can perform homogeneous reconstruction with a random initialization of the volume, avoiding the need for prior information about the molecule. Although it demonstrated the possibility of using a non-variational encoder to predict the orientations $\rot$, cryoPoseNet assumes the translations $\trans$ to be given and the volume is stored in real space in the decoder (while the image formation model is in Fourier space, see Sec.~\ref{subsec:imf}), thereby requiring a 3D Fourier transform at each forward pass and making the overall decoding step slow. The volume reconstructed by cryoPoseNet often gets stuck in local minima, which is a problem we also address in this paper (see Sec.~\ref{subsec:loss}). Finally, the two last methods only proved they could be used with simulated datasets and, to the best of our knowledge, no amortized inference technique for volume estimation from unknown poses have been proven to work with experimental datasets in cryo-EM.\\

Previous methods differ in the way poses are inferred in the generative model. Yet, the only variable of interest is the description of the conformational state (for heterogeneous methods) and associated molecular volumes, while poses can be considered ``nuisance'' variables. As a result, recent works have explored methods that avoid the inference of poses altogether, such as GAN-based approaches~\cite{akccakaya2022unsupervised}. CryoGAN \cite{gupta_cryogan_2021}, for example, used a cryo-EM simulator and a discriminator neural network to optimize a 3D volume. Although preliminary results are shown on experimental datasets, the reconstruction cannot be further refined with other methods due to the absence of predicted poses.

Our approach performs an amortized inference of poses and therefore circumvents the need for expensive searches over $SO(3)\times\mathbb{R}^2$, as in non-amortized techniques. In the implementation, no parameter needs to be statically associated with each image. Consequently, the memory footprint and the runtime of our algorithm does not scale with the number of images in the dataset. We introduce a loss function called ``symmetric loss'' that prevents the model from getting stuck in local minima with spurious planar symmetries. Finally, in contrast to previous amortized inference techniques, our method can perform volume reconstruction on experimental datasets.

\section{Methods}\label{sec:methods}

\subsection{Image Formation Model and Fourier-slice Theorem}
\label{subsec:imf}

\begin{figure}[t!]
\centering
\includegraphics[width=\textwidth]{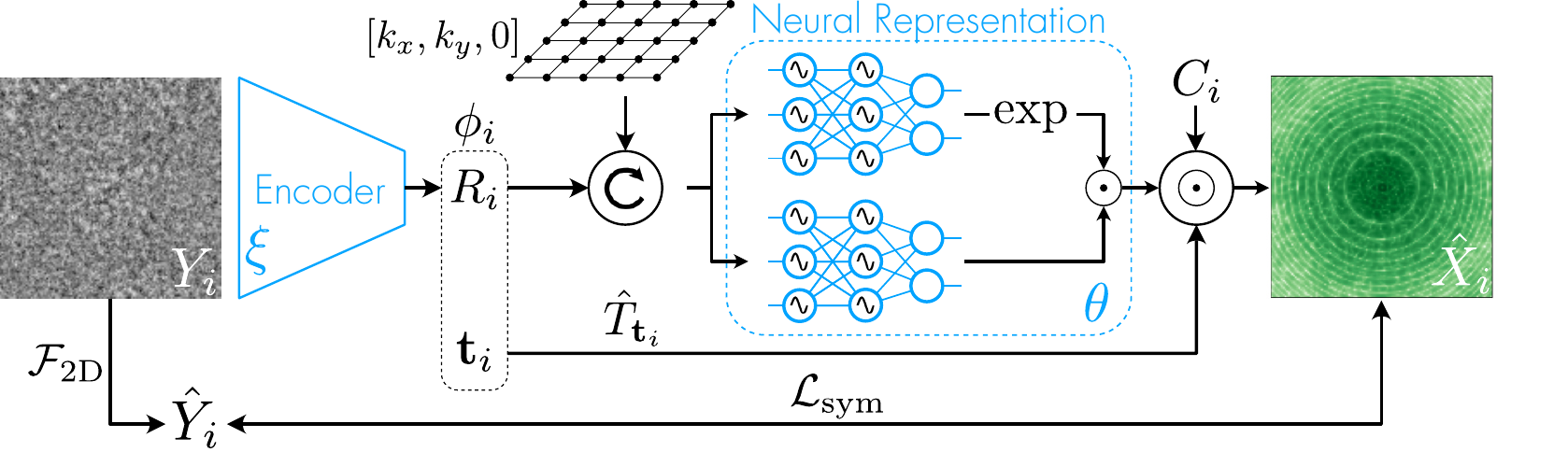}
\caption{Overview of our pipeline. The encoder, parameterized by $\enc$ learns to map images $\projgt$ to their associated pose $\pose=(\rot,\trans)$. The matrix $R_i$ rotates a slice of 3D coordinates in Fourier space. The coordinates are fed into a neural representation of $\fvol$, parameterized by $\theta$. The output is multiplied by the CTF $\ctf$ and the translation operator $\hat{T}_{\trans}$ to build $\fprojpred$, a noise-free estimation of $\Ftwo[\projgt]=\fprojgt$. $\fprojpred$ and $\fprojgt$ are compared \textit{via} the symmetric loss $\mathcal{L}_{\textrm{sym}}$. Differentiable parameters are represented in blue.}
\label{fig:overview}
\end{figure}

In a cryo-EM sample, the charges carried by each molecule and their surrounding environment create an electrostatic potential that scatters probing electrons, which we refer to as the electron scattering ``volume,'' and consider as a mapping
\begin{equation}
    \vol: \mathbb{R}^3 \to \mathbb{R}.
\end{equation}
In the sample, each molecule $i$ is in an unknown orientation $\rot\in SO(3) \subset \mathbb{R}^{3\times 3}$. The probing electron beam interacts with the electrostatic potential and its projections
\begin{equation}
    \surf: (x,y) \mapsto \int_z \vol\left(\rot\cdot\left[x, y, z\right]^T\right)dz
\label{eq:proj}
\end{equation}
are considered mappings from $\mathbb{R}^2$ to $\mathbb{R}$. The beam then interacts with the lens system characterized by the Point Spread Function (PSF) $\psf$ and individual particles are cropped from the full micrograph. The obtained images may not be perfectly centered on the molecule and small translations are modeled by $\trans\in\mathbb{R}^2$. Finally, taking into account signal arising from the vitreous ice into which the molecules are embedded as well as the non-idealities of the lens and the detector, each image $\projgt$ is generally modeled as
\begin{equation}
    \projgt = T_{\trans} * \psf * \surf + \noise
\label{eq:ifmreal}
\end{equation}
where $*$ is the convolution operator, $T_\mathbf{t}$ the $\mathbf{t}$-translation kernel and $\noise$ white Gaussian noise on $\mathbb{R}^2$ \cite{vulovic_image_2013,scheres_relion_2012}.

With a formulation in real space, both the integral over $z$ in Eq.~\eqref{eq:proj} and the convolution in Eq.~\eqref{eq:ifmreal} make the simulation of the image formation model computationally expensive. A way to avoid these operations is to use the Fourier-slice Theorem~\cite{bracewell1956strip}, which states that for any volume $\vol$ and any orientation $\rot$,
\begin{equation}
    \Ftwo\left[\surf\right] = \slice\left[\Fthree\left[\vol\right]\right],
\end{equation}
where $\Ftwo$ and $\Fthree$ are the 2D and 3D Fourier transform operators and $\slice$ the ``slice'' operator defined such that for any $\fvol:\mathbb{R}^3\to\mathbb{C}$,
\begin{equation}
    \slice[\fvol]:(k_x, k_y) \mapsto \fvol\left(\rot\cdot[k_x,k_y,0]^T\right).
\label{eq:slicing}
\end{equation}
That is, $\slice[\fvol]$ corresponds to a 2D slice of $\fvol$ with orientation $\rot$ and passing through the origin. In a nutshell, if $\fprojgt=\Ftwo[\projgt]$ and $\fvol=\Fthree\left[\vol\right]$, the image formation model in Fourier space can be expressed as
\begin{equation}
    \fprojgt = \hat{T}_{\trans} \odot \ctf \odot \slice[\fvol] + \fnoise,
\label{eq:imffourier}
\end{equation}
where $\odot$ is the element-wise multiplication, $\ctf= \Ftwo\left[\psf\right]$ is the Contrast Transfer Function (CTF), $\hat{T}_\mathbf{t}$ the $\mathbf{t}$-translation operator in Fourier space (phase shift) and $\fnoise$ complex white Gaussian noise on $\mathbb{R}^2$. Based on this generative model, \cryoSIMBA solves the inverse problem of inferring $\fvol$, $\rot$ and $\trans$ from $\fprojgt$ assuming $\ctf$ is known.

\subsection{Overview of \CryoSIMBA}

\CryoSIMBA is built with an autoencoder architecture (see Fig.~\ref{fig:overview}). The encoder takes an image $\projgt$ as input and outputs a predicted orientation $\rot$ along with a predicted translation $\trans$ (Sec.~\ref{subsec:PE}). $\rot$ is used to rotate a 2-dimensional grid of $\sidelen^2$ 3D-coordinates $[k_x, k_y, 0]\in\mathbb{R}^3$ which are then fed into the neural network $\fvol_\theta$. This neural network is an implicit representation of the current estimate of the volume $\fvol$ (in Fourier space), and this query operation corresponds to the ``slicing'' defined by Eq.~\eqref{eq:slicing} (Sec.~\ref{subsec:FN}). Based on the estimated translation $\trans$ and given CTF parameters $\ctf$, the rest of the image formation model described in Eq.~\eqref{eq:imffourier} is simulated to obtain $\fprojpred$, a noise-free estimation of $\fprojgt$. These images are compared using a loss described in Sec.~\ref{subsec:loss} and gradients are backpropagated throughout the differentiable model in order to optimize both the encoder and the neural representation.

\subsection{Pose Estimation}
\label{subsec:PE}

\CryoSIMBA uses a Convolutional Neural Network (CNN) to predict the parameters $\rot$ and $\trans$ from a given image, thereby avoiding expensive orientation matching computations performed by other methods \cite{scheres_relion_2012,punjani_cryosparc_2017,zhong_cryodrgn2_2021}. The architecture of this encoder has three layers.
\begin{enumerate}
    \item \textit{Low-pass filtering}: $\projgt\in\mathbb{R}^{\sidelen\times\sidelen}$ is fed into a bank of Gaussian low-pass filters.
    \item \textit{Feature extraction}: the filtered images are stacked channel-wise and fed into a CNN whose architecture is inspired by the first layers of VGG16~\cite{simonyan2014very}, which is known to perform well on image classification tasks.
    \item \textit{Pose estimation}: this feature vector finally becomes the input of two separate fully-connected neural networks. The first one outputs a vector of dimension $6$ of $S^2\times S^2$ \cite{zhou_continuity_2020} (two vectors on the unitary sphere in $\mathbb{R}^3$) and converted into a matrix $\rot\in\mathbb{R}^{3\times 3}$ using the PyTorch3D library \cite{ravi_accelerating_2020}. The second one outputs a vector of dimension $2$, directly interpreted as a translation vector $\trans \in \mathbb{R}^2$.
\end{enumerate}
We call $\enc$ the set of differentiable parameters in the encoder described above. We point the reader to Supp.~B for more details about the architecture of the encoder.

\subsection{Neural Representation in Fourier Space (FourierNet)}
\label{subsec:FN}

Instead of using a voxel-based representation, we maintain the current estimate of the volume using a neural representation. This representation is parameterized by $\theta$ and can be see seen as a mapping $\fvol_\theta:\mathbb{R}^3\to\mathbb{C}$.

In imaging and volume rendering, neural representations have been used to approximate signals defined in real space \cite{park2019deepsdf,atzmon2020sal,genova2019learning,michalkiewicz2019implicit,sitzmann2019scene}. Neural Radiance Field (NeRF) \cite{mildenhall2020nerf} is a successful technique to maintain a volumetric representation of a real scene. A view-independent NeRF model, for example, maps real 3D-coordinates $[x,y,z]$ to a color vector and a density scalar using positional encoding \cite{vaswani2017attention} and a set of fully-connected layers with ReLU activation functions. Sinusoidal Representation Networks (SIRENs) \cite{sitzmann2020implicit} can also successfully approximate 3D signed distance functions with a shallow fully-connected neural network using sinusoidal activation functions. However, these representations are tailored to approximate signals defined in real space. Here, we want to directly represent the Fourier transform of the electrostatic potential of a molecule. Since this potential is a smooth function of the spatial coordinates, the amplitude of its Fourier coefficients $\fvol(\mathbf{k})$ is expected to decrease with $|\mathbf{k}|$, following a power law (see Supp.~C for more details). In practice, this implies that $|\fvol|$ can vary over several orders of magnitude and SIRENs, for example, are known to poorly approximate these types of functions~\cite{sitzmann2020implicit}. The first method to use neural representations for volume reconstruction in cryo-EM, cryoDRGN~\cite{zhong_reconstructing_2019,zhong_cryodrgn2_2021}, proposed to use a Multi-Layer Perceptron (MLP) with positional encoding in Hartley space (where the FST still applies).

With our work, we introduce a new kind of neural representation (FourierNet), tailored to represent signals defined in the Fourier domain, inspired by the success of SIRENs for signals defined in real space. Our idea is to allow a SIREN to represent a signal with a high dynamic range by raising its output in an exponential function. Said differently, the SIREN only represents a signal that scales logarithmically with the approximated function. Since Fourier coefficients are defined on the complex plane, we use a second network in our implicit representation to account for the phase variations. This architecture is summarized in Fig.~\ref{fig:overview} and details on memory requirements are given in Supp.~C. Input coordinates $[k_x, k_y, k_z]$ are fed into two separate SIRENs outputting 2-dimensional vectors. For one of them, the exponential function is applied element-wise and the two obtained vectors are finally element-wise mutliplied to produce a vector in $\mathbb{R}^2$, mapped to $\mathbb{C}$ with the Cartesian coordinate system. Since $\fvol_\theta$ must represent the Fourier transform of real signals, we know that it should verify $\fvol_\theta(-\mathbf{k}) = \fvol_\theta(\mathbf{k})^*$. We enforce this property by defining
\begin{equation}
    \fvol_\theta(\mathbf{k}) = \fvol_\theta(-\mathbf{k})^* \quad \text{if }k_x<0.
\end{equation}
Benefits of this neural representation are shown on 2-dimensional signals in the Supp.~C.

The neural representation is queried for a set of $\sidelen^2$ 3D-coordinates $[k_x,k_y,k_z]$, thereby producing a discretized slice $\slice[\fvol_\theta]\in\mathbb{C}^{\sidelen\times\sidelen}$. The rest of the image formation model \eqref{eq:imffourier} is simulated by element-wise multiplying $\slice[\fvol_\theta]$ by the CTF $\ctf$ and a translation matrix,
\begin{equation}
    \fprojpred = \hat{T}_{\trans} \odot \ctf \odot \slice[\fvol_\theta],
\end{equation}
where $\hat{T}_{\trans}$ is defined by
\begin{equation}
    \hat{T}_{\trans}(\mathbf{k}) = \exp\left(-2j\pi\mathbf{k}\cdot\trans\right).
\end{equation}
The parameters of the CTF are provided by external CTF estimation softwares such as CTFFIND~\cite{rohou2015ctffind4}. The whole encoder--decoder pipeline can be seen as a function that we call $\simba$, such that $\fprojpred = \simba(\projgt)$.

\subsection{Symmetric Loss}
\label{subsec:loss}

In the image formation model of Eq.~\eqref{eq:ifmreal}, the additive noise $\noise$ is assumed to be Gaussian and uncorrelated (white Gaussian noise) \cite{vulovic_image_2013,scheres_relion_2012}, which means that its Fourier transform $\fnoise$ follows the same kind of distribution. Therefore, maximum likelihood estimation on a batch $\batch$ amounts to the minimization of the L2-loss.

Nonetheless, we empirically observed that using this loss often led the model to get stuck in local minima where the estimated volume showed spurious planar symmetries (see Sec.~\ref{subsec:SL}). We hypothesize that this behaviour is linked to the fundamental ambiguity contained in the image formation model in which, given unknown poses, one cannot distinguish two ``mirrored'' versions of the same volume~\cite{rosenthal2003optimal}. We discuss this hypothesis in more detail in Supp.~D. To solve this problem, we designed a loss that we call ``symmetric loss'' defined as
\begin{equation}
    \symloss = \sum_{i\in\batch} \text{min}\,\big\{ \lVert \fprojgt - \simba(\projgt) \rVert^2, \lVert \flip[\fprojgt] - \simba\left(\flip\left[\projgt\right]\right) \rVert^2\big\}
\label{eq:symloss}
\end{equation}
where $\flip$ applies an in-plane rotation of $\pi$ on $\sidelen\times\sidelen$ images. Using the symmetric loss, the model can be supervised on a set of images $\projgt$ in which the predicted in-plane rotation (embedded in the predicted matrix $\rot$) can always fall in $\left[-\pi/2,\pi/2\right]$ instead of $\left[-\pi,\pi\right]$. As shown in Sec. \ref{subsec:SL} and explained in Supp.~D, this prevents \cryoSIMBA from getting stuck in spuriously symmetrical states.

\section{Results}\label{sec:results}

We qualitatively and quantitatively evaluate \cryoSIMBA for \emph{ab initio} reconstruction of both simulated and experimental datasets. We first compare \cryoSIMBA to the state-of-the-art method cryoSPARC \cite{punjani_cryosparc_2017} in terms of runtime on a simulated dataset of the \emph{80S} ribosome with low levels of noise. We then compare our method with baseline methods in terms of resolution and pose accuracy on simulated datasets with and without noise (\emph{spike}, \emph{spliceosome}). Next, we show that \cryoSIMBA can perform \emph{ab initio} reconstruction on an experimental cryo-EM dataset (\emph{80S}), which is the first time for a method estimating poses in an amortized fashion. Finally, we highlight the importance of a tailored neural representation in the decoder and the role of the symmetric loss in an ablation study.

\subsection{Reconstruction on Simulated Datasets}

\begin{figure}[t]
\centering
\includegraphics[width=\textwidth]{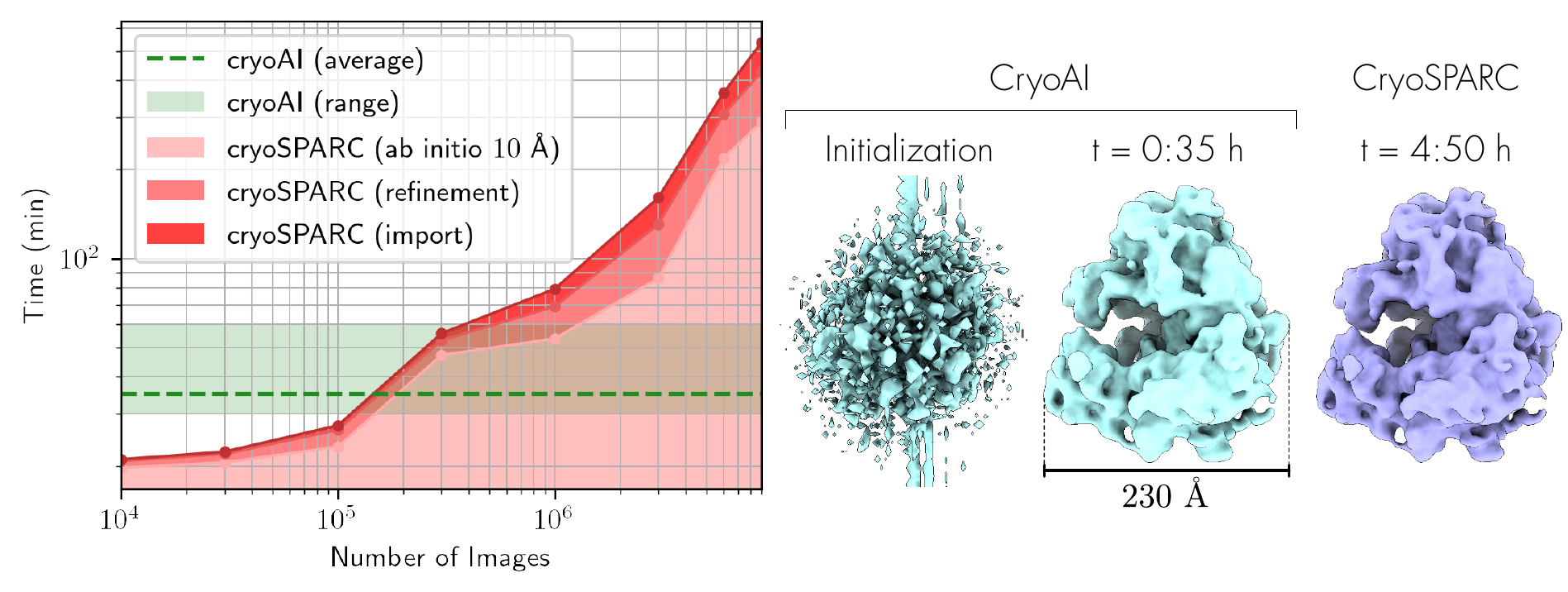}
\caption{(Left) Time to reach $10$~\r{A} of resolution with \cryoSIMBA (range and average over $5$ runs per datapoint) and cryoSPARC vs. number of images in the simulated 80S dataset. (Right) Estimated volume at initialization and after $35~\text{min}$ of running \cryoSIMBA vs. cryoSPARC after convergence, with 9M images.}
\label{fig:timeplot}
\end{figure}

\textbf{Experimental Setup.} We synthesize three datasets from deposited Protein Data Bank (PDB) structures of the Plasmodium falciparum 80S ribosome (PDB: 3J79 and 3J7A)~\cite{wong2014cryo}, the SARS-CoV-2 spike protein (PDB: 6VYB) \cite{walls2020structure} and the pre-catalytic spliceosome (PDB: 5NRL) \cite{plaschka2017structure}. First, a 3D grid map, the \emph{ground-truth volume}, is generated in ChimeraX~\cite{goddard2018ucsf} from each atomic model using the steps described in Supp.~A. Then a dataset is generated from the ground-truth volume using the image formation model described in Sec.~\ref{subsec:imf}. Images are sampled at $\sidelen=128$. Rotations $\rot$ are randomly generated following a uniform distribution over $SO(3)$ and random translations $\trans$ are generated following a zero-mean Gaussian distribution ($\sigma=20$~\r{A}). The defocus parameters of the CTFs are generated with a log-normal distribution. We build noise-free (ideal) and noisy versions of each dataset ($\text{SNR}=0\text{dB}$ for 80S, $\text{SNR}=-10\text{dB}$ for the others, see Supp.~E for details). We compare \cryoSIMBA with three baselines: the state-of-the-art software cryoSPARC v3.2.0~\cite{punjani_cryosparc_2017} with default settings, the neural network--based method cryoDRGN2 \cite{zhong_cryodrgn2_2021} and the autoencoder-based method cryoPoseNet \cite{nashed_cryoposenet_2021} (with the image formation model in real space in the decoder, see Supp.~A). We quantify the accuracy of the reconstructed volume by computing the Fourier Shell Correlations (FSC) between the reconstruction and the ground truth and reporting the resolution at the $0.5$ cutoff. All experiments are run on a single Tesla V100 GPU with 8 CPUs.

\textbf{Convergence Time.} We compare \cryoSIMBA with cryoSPARC in terms of runtime for datasets of increasing size in Fig.~\ref{fig:timeplot}. We use the simulated 80S dataset and define the running time as the time needed to reach a resolution of $10$~\r{A} ($2.65~\text{pixels}$), which is a sufficiently accurate resolution to perform refinement with cryoSPARC (see workflow in Supp.~A). With default parameters, cryoSPARC's ab initio reconstruction must process all images in the dataset. We show the time required by cryoSPARC for importing data and for the refinement step. \CryoSIMBA processes images batch-wise and does not statically associates variables to each image, making the convergence time (for reaching the specified resolution) independent from the size of the dataset. By contrast, the computation time of cryoSPARC increases with the number of images and can reach $5$~hours with a dataset of $9\text{M}$ particles. We additionally show in Supp~F the time required to estimate all the poses of the dataset with \cryoSIMBA's encoder.

\textbf{Accuracy.} We compare \cryoSIMBA with baseline methods on the \textit{spike} and \textit{spliceosome} datasets in Table~\ref{table:simres}. We compare the reconstructed variables (volume and poses) with their ground truth values (from simulation). Results of cryoDRGN2 are reported from available data in \cite{zhong_cryodrgn_2021}. Images were centered for cryoPoseNet since the method does not predict $\trans$. A ``tight'' adaptive mask was used with cryoSPARC. The performance of \cryoSIMBA is comparable with the baselines. The splicesome and the noise-free spike protein are reconstructed with state-of-the-art accuracy. In the noisy spike dataset, the accuracy of \cryoSIMBA and cryoSPARC decreases,  which may be due to the pseudo-symmetries shown by the molecule (visual reconstruction in Supp.~F). CryoPoseNet gets stuck for at least 24 hours in a state where the the resolution is very poor on both spike datasets.

\begin{table}[t]
    \scriptsize
    \centering
    \caption{Accuracy of pose and volume estimation for simulated data. Resolution (Res.) is reported using the $\text{FSC}=0.5$ criterion, in pixels $(\downarrow)$. Rotation (Rot.) error is the median square Frobenius norm between predicted and ground truth matrices $\rot$ $(\downarrow)$. Translation (Trans.) error is the mean square L2-norm, in pixels $(\downarrow)$.}
    \ra{1.2}
    \begin{tabular}{llcccccccc}\toprule
    \multicolumn{2}{c}{Dataset} & \phantom{abc} & cryoPoseNet & \phantom{abc} & cryoSPARC & \phantom{abc} & cryoDRGN2 & \phantom{abc} & \cryoSIMBA\\
    \midrule
    \textit{Spliceosome (ideal)} &
    Res. && $2.78$ && $\mathbf{2.13}$ && --- && $\mathbf{2.13}$ \\
    & Rot. && $0.004$ && $\mathbf{0.0002}$ && --- && $0.0004$ \\
    & Trans. && --- && $0.006$ && --- && $\mathbf{0.001}$ \\
    \midrule
    \textit{Spliceosome (noisy)} &
    Res. && $3.15$ && $\mathbf{2.61}$ && --- && $\mathbf{2.61}$\\
    & Rot. && $0.01$ && $\mathbf{0.002}$ && --- && $0.007$ \\
    & Trans. && --- && $\mathbf{0.007}$ && --- && $0.01$ \\
    \midrule
    \textit{Spike (ideal)} &
    Res. && 16.0 && $2.33$ && --- && $\mathbf{2.29}$\\
    & Rot. && $5$ && $0.0003$ && $\mathbf{0.0001}$ && $0.0003$  \\
    & Trans. && --- && $0.007$ && --- && $\mathbf{0.001}$  \\
    \midrule
    \textit{Spike (noisy)} & 
    Res. && 16.0 && $3.56$ && $\mathbf{2.03}$ && $2.91$\\
    & Rot. && $6$ && $0.02$ && $\mathbf{0.01}$ && $\mathbf{0.01}$ \\
    & Trans. && --- && $0.008$ && --- && $\mathbf{0.003}$ \\
    \bottomrule
    \end{tabular}
    \label{table:simres}
\end{table}

\subsection{Reconstruction on Experimental Datasets}

\begin{figure}[t]
\centering
\includegraphics[width=\textwidth]{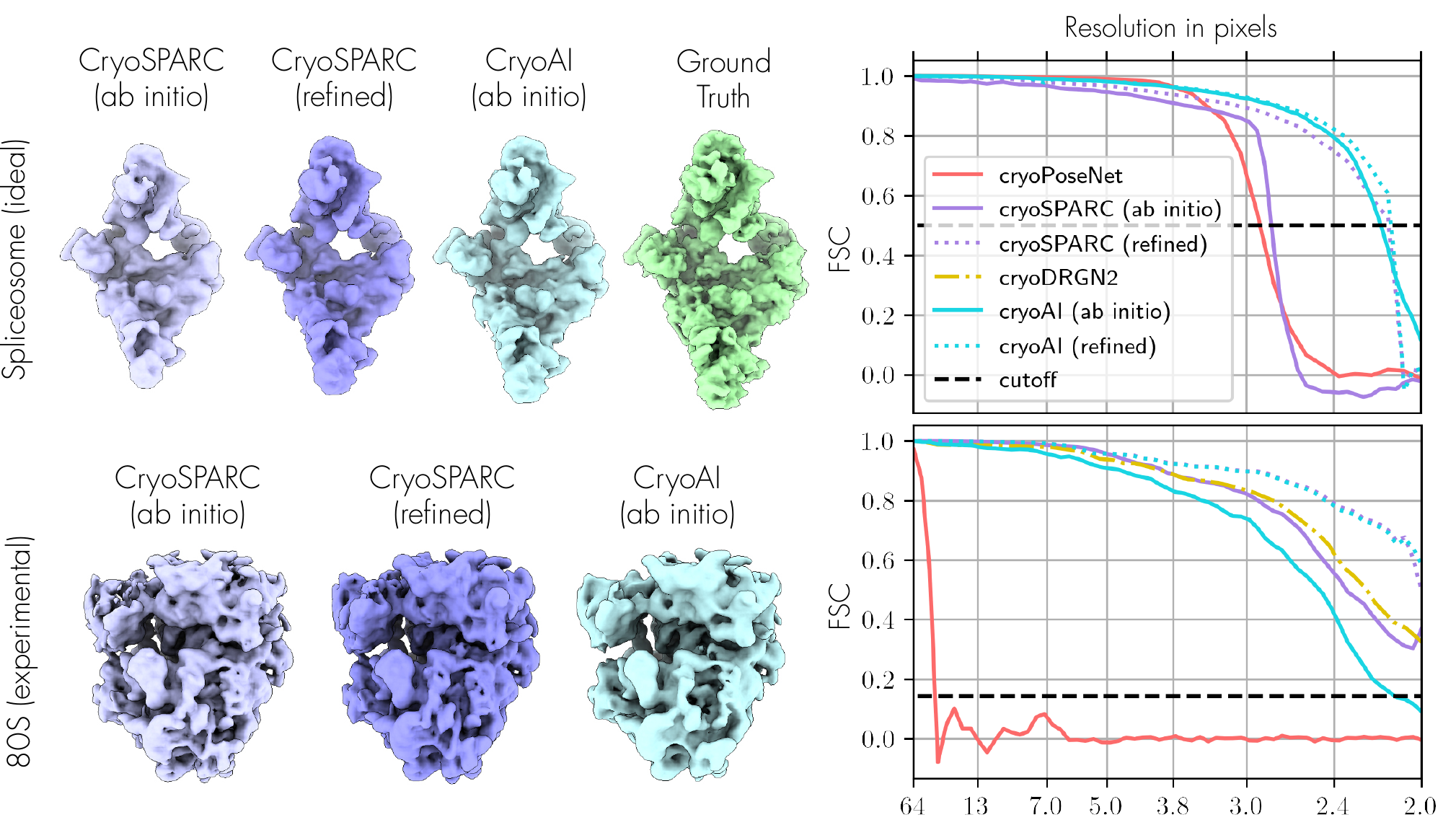}
\caption{(Top left) Volume reconstruction on a noise-free simulated dataset of the spliceosome ($\sidelen=128$, pixel size $=4.25$~\r{A}). (Bottom left) Volume reconstruction for the experimental 80S dataset ($\sidelen=128$, pixel size $=3.77$~\r{A}). (Right) Fourier Shell Correlations, reconstruction-to-ground-truth (top) or reconstruction-to-reconstruction (bottom). A resolution of $2.0$~pixels corresponds to the Nyquist frequency. \CryoSIMBA can be refined using the software cryoSPARC.}
\label{fig:expresribo}
\end{figure}

\textbf{Experimental Setup.} We use the publicly available 80S experimental dataset EMPIAR-10028 \cite{wong2014cryo,cryodrgn_empiar,iudin2016} containing $105{\small,}247$ images of length $\sidelen=360$ ($1.34$~\r{A} per pixel), downsampled to $\sidelen=256$. The dataset is evenly split in two, each method runs independent reconstructions on each half and the FSC are measured between the two reconstructions. We compare \cryoSIMBA with cryoPoseNet and cryoSPARC. The dataset fed to \cryoSIMBA and cryoPoseNet is masked with a circular mask of radius $84$~pixels, while cryoSPARC adaptively updates a ``tight" mask. \CryoSIMBA and cryoPoseNet reconstruct a volume of size $128^3$. For cryoSPARC, both the \textit{ab initio} volume and the volume subsequently homogeneously refined from it were downsampled to the same size $128^3$.
We also demonstrate the possibility of refining \cryoSIMBA's output with the software cryoSPARC.
Finally, we report the results published for cryoDRGN2 \cite{zhong_cryodrgn2_2021} that were obtained on a filtered version of the same dataset \cite{cryodrgn_empiar} downsampled to $\sidelen=128$ prior reconstruction.

\textbf{Results.} We report quantitative and qualitative results in Fig.~\ref{fig:expresribo}. \CryoSIMBA is the first amortized method to demonstrate proper volume reconstruction on an experimental dataset, although techniques predicting poses with an orientation-matching step (like cryoDRGN2) or followed by an EM-based refinement step (like cryoSPARC) can reach slightly higher resolutions. State-of-the-art results can be obtained with cryoSPARC's refinement, initialized from either cryoSPARC's or \cryoSIMBA's \emph{ab initio}. Since simulated datasets were built using the same image formation model as the one \cryoSIMBA uses in its decoder, the gap in performance between the experimental and simulated datasets suggests that improvements could potentially be achieved with a more accurate physics model.

\subsection{Ablation Study}
\label{subsec:SL}

\begin{figure}[t]
\centering
\includegraphics[width=\textwidth]{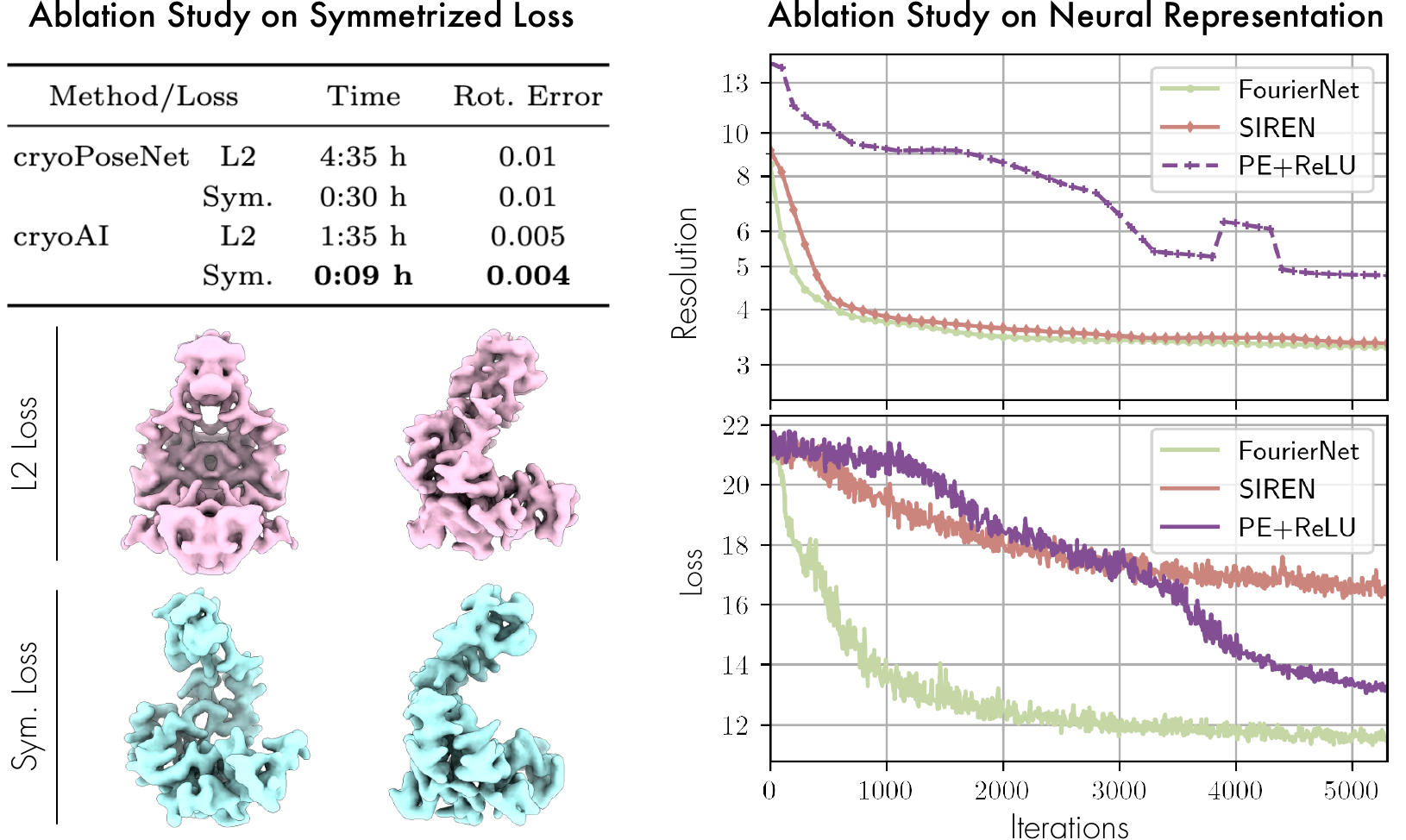}
\caption{(Top left) Ablation study on the symmetric loss with \cryoSIMBA and cryoPoseNet with simulated noise-free adenylate kinase ($\sidelen=64$). We report the minimal convergence time out of 5 runs. CryoPoseNet is always slower and achieves worse results. The symmetric loss always accelerates convergence. (Bottom left) Volume reconstruction when using a L2 loss vs. the symmetric loss. The latter prevents the model from getting stuck in a symmetrical local minimum. (Right) Loss and resolution (in pixels, $\text{FSC}=0.143$ cutoff) vs. number of iterations with a FourierNet, a SIREN \cite{sitzmann2020implicit} and an MLP with ReLU activation functions and positional encoding ($32$ images per batch).}
\label{fig:ablstudy}
\end{figure}

\textbf{Importance of Symmetric Loss.} The purpose of the symmetric loss is to prevent the model from getting stuck in local minima where the volume shows incorrect planar symmetries. Ullrich \textit{et al.} showed in \cite{ullrich_differentiable_2019} that optimizing the poses using a gradient-based method often leads the model to fall in sub-optimal minima, due to the high non-convexity of the optimization problem. In \cite{zhong_cryodrgn2_2021}, Zhong \textit{et al.} implemented an autoencoder-based method (dubbed PoseVAE), and compared it to cryoDRGN2. The method is unable to properly reconstruct a synthetic hand, and a spurious planar symmetry appears in their reconstruction. We use a noisy dataset ($\sidelen=128$) generated from a structure of Adenylate kinase (PDB~4AKE) \cite{muller1996adenylate}. We show in Fig.~\ref{fig:ablstudy} that our method presents the same kind of artifact when using a L2 loss and validate that the symmetric loss prevents these artifacts. In Fig.~\ref{fig:ablstudy}, we compare our method to cryoPoseNet with and without the symmetric loss on a simulated ideal dataset of the same molecule ($\sidelen=64$). Both methods use an autoencoder-based architecture and both converge significantly faster with the symmetric loss. With the same loss, \cryoSIMBA is always faster than cryoPoseNet since our method operates in Fourier space and avoids the approximation of integrals using the FST.

\textbf{Comparison of Neural Representations.} We replaced FourierNet with other neural representations in the decoder and compared the convergence rate of these models on the noisy Adenylate kinase dataset (\sidelen=128). In Fig.~\ref{fig:ablstudy}, we compare our architecture with a  multi-layer perceptron (MLP) with sinusoidal activation functions (i.e., a SIREN \cite{sitzmann2020implicit}) and an MLP with ReLU activation function and positional encoding, as used by cryoDRGN2 \cite{zhong_cryodrgn2_2021}. We keep approximately $300\text{k}$ differentiable parameters in all representations. FourierNet significantly outperforms the two other architectures in terms of convergence speed.

\section{Discussion}\label{sec:discussion}

The amount of collected cryo-EM data is rapidly growing~\cite{namba_2022}, which increases the need for efficient \textit{ab initio} reconstruction methods.
\CryoSIMBA proposes a framework of amortized inference to meet this need by having a complexity that does not grow with the size of the dataset. Since \CryoSIMBA jointly estimates volume and poses, it can be followed by reconstruction methods that address conformational heterogeneities, such as the ones available in cryoSPARC \cite{punjani_cryosparc_2017}, RELION \cite{scheres_relion_2012}, or cryoDRGN \cite{zhong_cryodrgn_2021}. The ever increasing size of cryo-EM datasets is necessary to provide sufficient sampling of conformational heterogeneities with increasing accuracy, in particular when imaging molecules that display complex dynamics. However, existing methods that tackle the more complex inference task of heterogeneous reconstruction also see their runtime suffer as datasets grow bigger, again showing the need for new developments that leverage amortized inference.

Future work on \cryoSIMBA includes adding features to the image formation model implemented in the decoder. CTFs, for example, are currently only characterized by three parameters (two defoci parameters and an astigmatism angle) but could be readily enhanced to account for higher-order effects (see e.g. \cite{zivanov2018new}). A richer noise model, currently assumed to be Gaussian and white, could also improve the performance of the algorithm. In order to tackle the case of very noisy experimental datasets, adaptive masking techniques, such as those used by cryoSPARC, could be beneficial. In terms of hardware development, \cryoSIMBA would benefit from being able to run on more than a single GPU using data parallelism and/or model parallelism, thereby improving both runtime and efficiency. \CryoSIMBA, as described here, belongs to the class of homogeneous reconstruction methods; future developments should explore its performance in an heterogenous reconstruction setting, where conformational heterogeneity is baked in the generative model and the encoder is enhanced to predict descriptions of conformational states in low-dimensional latent space along with the poses.

\textbf{Conclusion.} Advancing our understanding of the building blocks of life hinges upon our ability to leverage cryo-EM at its full potential. While recent advances in instrumentation and hardware have enabled massive datasets to be recorded at unprecedented throughput, advancing the associated algorithms to efficiently scale with these datasets is crucial for the field to move forward. Our work presents important steps towards this goal.

\textbf{Acknowledgment.} We thank Wah Chiu for numerous discussions that helped shape this project. This work was supported by the U.S. Department of Energy, under DOE Contract No. DE-AC02-76SF00515. N.M. acknowledges support from the National Institutes of Health (NIH), grant No. 1R01GM144965-01. We acknowledge the use of the computational resources at the SLAC Shared Scientific Data Facility (SDF).

%
%
\bibliographystyle{splncs04}
\bibliography{main}

\end{document}


\pagestyle{headings}
\mainmatter
\def\ECCVSubNumber{3418}  

\title{Supplementary Material\\\CryoSIMBA: Amortized Inference of Poses for Ab Initio Reconstruction of 3D Molecular Volumes from Real Cryo-EM Images} 

\titlerunning{CryoAI}
%
\author{
Axel Levy$^{1,2,*}$\and
Fr\'{e}d\'{e}ric Poitevin$^{1,*}$\and
Julien Martel$^{2,*}$\and
Youssef Nashed$^{3}$\and
Ariana Peck$^{1}$\and
Nina Miolane$^{4}$\and
Daniel Ratner$^{3}$\and
Mike Dunne$^{1}$\and
Gordon Wetzstein$^{2}$
}
\authorrunning{A. Levy, F. Poitevin, J. Martel et al.}
%
\institute{
$^{1}$LCLS, SLAC National Accelerator Laboratory, Menlo Park, CA, USA\\
$^{2}$Stanford University, Department of Electrical Engineering, Stanford, CA, USA\\
$^{3}$ML Initiative, SLAC National Accelerator Laboratory, Menlo Park, CA, USA\\
$^{4}$University of California Santa Barbara, Department of Electrical and Computer Engineering, Santa Barbara, CA, USA
}

\maketitle

\section{Experimental Details}\label{supp:expsetup}
CryoAI and all the baselines are run on a single Tesla V100 GPU with 8 CPUs. Fourier Shell Correlations between two voxel grids (aligned with an exhaustive search over $SO(3)\times\mathbb{R}^3$) are measured using the software EMAN~v2.91~\cite{tang_eman2_2007}.

\subsection{Generation of Ground Truth Volume from Atomic Model}

The python library gemmi \cite{gemmi} was used to assemble a complete atomic model of the 80S Ribosome \cite{10.7554/eLife.03080} from its small (PDB:3J7A) and large (PDB:3J79) subunits deposited in different files in the Protein Data Bank (PDB) \cite{rcsbpdb}. The atomic models of the pre-catalytic Spliceosome (PDB:5NRL) \cite{spliceosome} and the SARS-CoV-2 spike ectodomain structure (PDB:6VYB) \cite{spikeprotein} were used without pre-processing after being downloaded from the PDB.

ChimeraX \cite{pettersen2021ucsf} was used to simulate a noise-less electrostatic potential volume discretized on a cubic grid from each atomic model. First, an empty cubic grid of desired spacing and extent is centered on the center of mass of the atomic model. Second, a map of same grid spacing is simulated from the atomic model and resampled on the first map, before being saved to a MRC file \cite{mrcformat}.

\subsection{\CryoSIMBA}

 Each minibatch contains $32$ images when $\sidelen=128$ and $128$ images when $\sidelen=64$. We use the Adam optimizer~\cite{kingma2014adam}, with a learning rate of $10^{-4}$.

\subsection{CryoSPARC}

CryoSPARC v3.2 \cite{punjani_cryosparc_2017} was used. We followed the typical workflow: \emph{Import} particle stacks, perform \emph{Ab Initio} reconstruction before \emph{Homogeneous Refinement}. Default parameter values were used except when noted in the text.

\subsection{CryoDRGN2~\cite{zhong_cryodrgn2_2021}}

All results from cryoDRGN2 are directly reported from the main paper or the supplementary materials, because source code for cryoDRGN2 is currently not available.

\subsection{CryoPoseNet~\cite{nashed_cryoposenet_2021}}

Volumes are represented with a voxel grid in real space and the image formation model is simulated in real space in the decoder. The encoder outputs rotations $\rot$ in the $6$-dimensional space $\mathcal{S}^2\times\mathcal{S}^2$. Each minibatch contains $32$ images when $\sidelen=128$ and $128$ images when $\sidelen=64$. The Adam optimizer~\cite{kingma2014adam} is used, with a learning rate of $10^{-3}$.

\section{Encoder Architecture}\label{supp:encoder}
\begin{figure}[h]
\centering
\includegraphics[width=\textwidth]{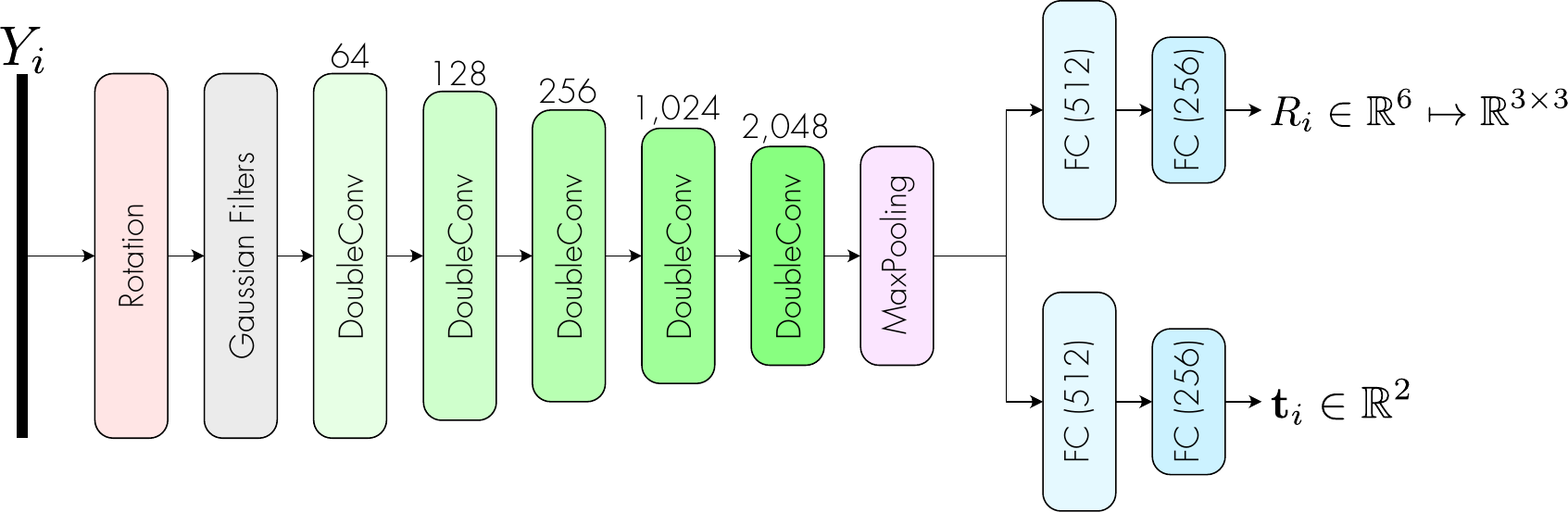}
\caption{Architecture of the encoder. The ``rotation'' layer duplicates each image and rotates one of the copies. ``Gaussian filters'' are low-pass filters. ``FC'' stands for ``fully connected''.}
\label{fig:encoderarchi}
\end{figure}

The encoder takes as input an image $\projgt$ and outputs a rotation matrix $\rot$ and a translation vector $\trans$. The rotation is represented in the 6-dimensional space $\mathcal{S}^2 \times \mathcal{S}^2$ \cite{zhou_continuity_2020} and converted in a matrix using the PyTorch3D library \cite{ravi_accelerating_2020}. The architecture of the encoder is summarized in Fig.~\ref{fig:encoderarchi}. Each image is first duplicated by a ``rotation'' layer, which applies an in-plane rotation of $\pi$ to one of the duplicates. Those duplicates are then concatenated batch-wise. Each image is filtered by a set of $5$ Gaussian filters of size $11$ with cutoff frequencies distributed geometrically between $0.1$ and $10~\text{pix}^{-1}$. Empirically, we found this multi-scale representation to lead to more robust convergence of our framework in some cases. This set of filtered images is then fed into five consecutive ``DoubleConv'' layers generating respectively $64$, $128$, $256$, $1024$ and $2048$ channels. A ``DoubleConv'' layer contains two convolutional layers with kernels of size $3$ and a max-pooling layer dividing the height and the width of each image by $2$. The last ``DoubleConv'' layer is followed by another max-pooling layer after which the dimension of each image is finally divided by $2^6=64$. The feature vector has, at this point, $8182$ dimensions when the input image has a size of $128^2$ pixels. This feature vector is finally fed into two separated fully connected neural networks with ReLU activation functions and two hidden layers of sizes $512$ and $256$.

\section{Implicit Representation in Fourier Domain}\label{supp:fouriernet}
\subsection{Electrostatic Potentials in Fourier Space}
\label{supp:fourierstats}

\begin{figure}[h]
\centering
\includegraphics[width=\textwidth]{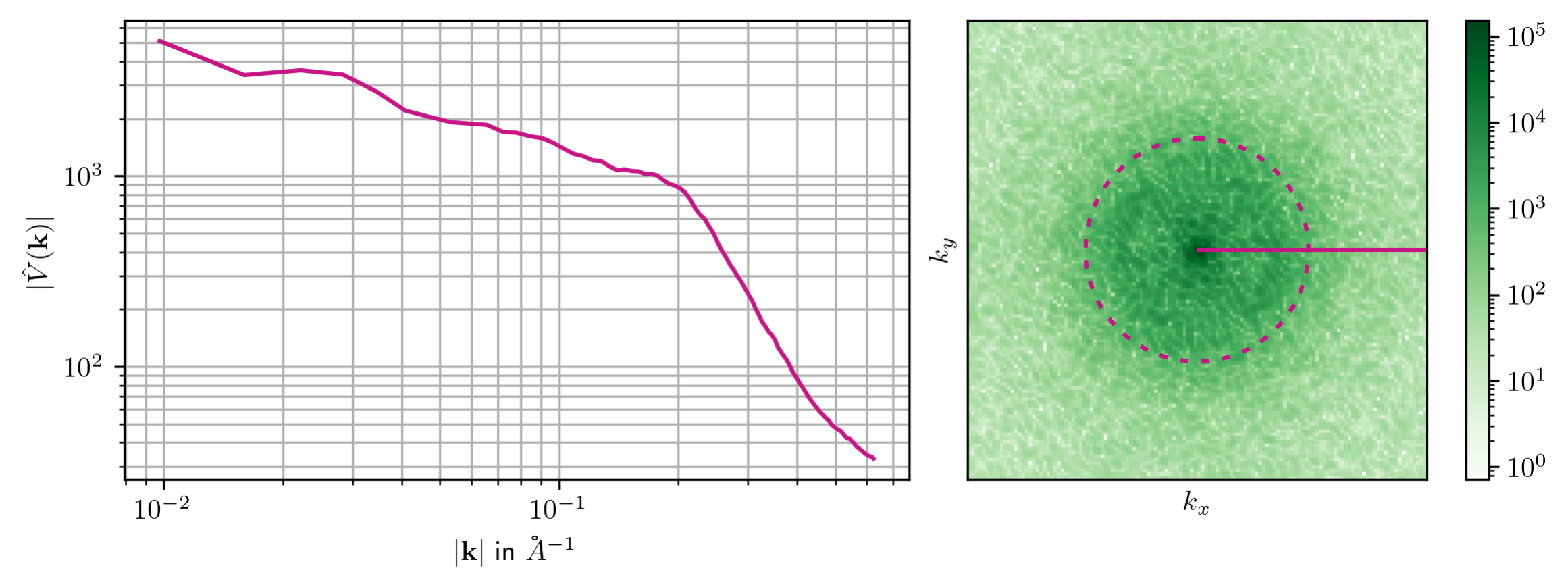}
\caption{(Left) $\fvol$ is the Fourier transform of the electron scattering potential of adenylate kinase ($128^3$ voxels of size $3.2$~\r{A}). The radial mean of $|\fvol|$ decreases following a power law of $|\mathbf{k}|$ and varies over 2 orders of magnitude. (Right) Slice $(k_x, k_y, 0)$ of $|\fvol|$, varying over $5$ orders of magnitude.}
\label{fig:fourierstats}
\end{figure}

The electrostatic potential $\vol(\mathbf{r})$ derives from the distribution of charges following the Poisson-Boltzmann equation \cite{pbreview}. We know that the smoothness of the function $V$ at low-to-medium resolution will translate into a rapid decrease of its Fourier coefficients $\fvol$. In particular, if we assume that $\vol$ is a smooth function of $\mathbf{r}$, in the sense that it is an $\alpha$-Lipschitz function, Sampson and Tuy \cite{sampson_fourier_1978} demonstrated that the norm of its Fourier transform $|\fvol(\mathbf{k})|$ decreases as least in $1/k^\beta$ for all $\beta<\alpha$. In Fig.~\ref{fig:fourierstats}, we analyze the Fourier transform of the electrostatic potential of an adenylate kinase molecule ($\sidelen=128$). We observe that $|\fvol(\mathbf{k})|$ indeed decreases rapidly with $|\mathbf{k}|$, which implies that $\fvol$ varies on several orders of magnitude.

\subsection{Performance of FourierNet on 2D images}
\label{supp:fouriernetperf}

\begin{figure}[t]
\centering
\includegraphics[width=0.8\textwidth]{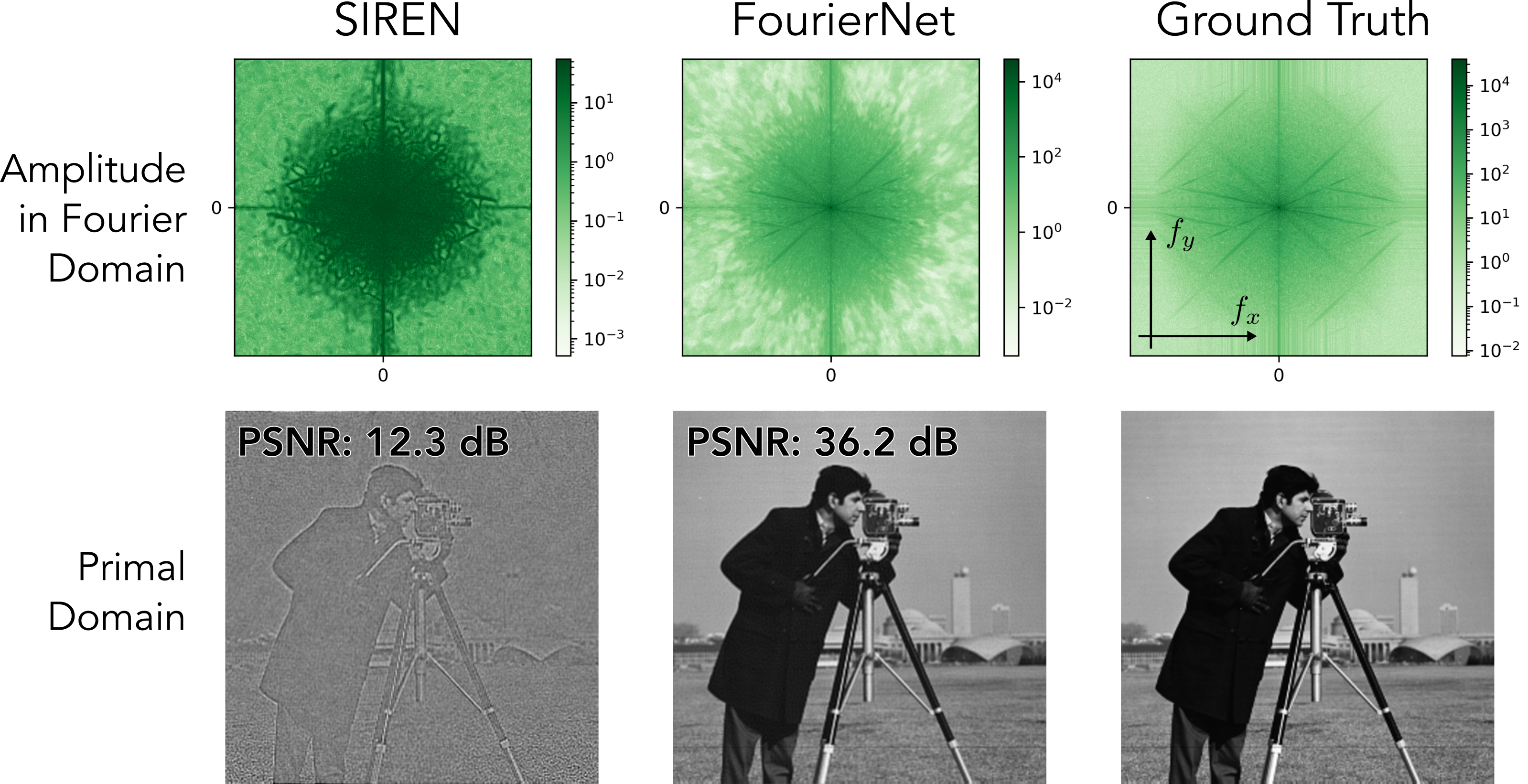}
\caption{Approximations of the Fourier transform of a ``natural'' image with a SIREN~\cite{sitzmann2020implicit} and a FourierNet, and their inverse Fourier transforms. The SIREN poorly fits signals that vary over several orders of magnitude.}
\label{fig:fouriernetperf}
\end{figure}

We show here that the architecture proposed in Fig.~2 is relevant for representing the Fourier transform of ``natural'' 2D signals by comparing it with a simple SIREN. Using gradient-based optimization we optimize the weights of a SIREN~\cite{sitzmann2020implicit} and of a FourierNet to approximate the Fourier transform of a real-world 2D image. For the comparison to be relevant, we use the same number of optimizable parameters ($300$k) in the SIREN and in the FourierNet. The results of the experiment are shown in Fig.~\ref{fig:fouriernetperf}. SIRENs are built in a way such that the weights follow a normalized distribution and they can only efficiently represent normalized functions. The ground truth Fourier transform varies on more than $6$ orders of magnitude, leading to a poor reconstruction when taking the inverse Fourier transform of the approximated function with a SIREN. On the opposite, the FourierNet can precisely fit the given Fourier transform, leading to a quantitatively better reconstruction in primal (real) domain. Artifacts, however, can still be observed at high frequency.

\subsection{Architecture Details}
\label{supp:fouriernetarchi}

Our neural representation uses two SIRENs with $256$ hidden features, mapping $\mathbb{R}^3$ to $\mathbb{R}^2$. The SIREN preceding the $\exp$ (see Fig.~2) has $2$ hidden layers while the other one contains $3$ hidden layers. The total number of parameters is only $330\small{,}240$, which can be compared to the number of parameters used in a voxel-based representation (slightly faster) of resolution $128$: $128^3=2\small{,}097\small{,}152$ (Table~\ref{table:repcomparison} also indicates the runtime on the \textit{kinase ideal} dataset).

\begin{table}[!h]
    \scriptsize
    \centering
    \caption{Comparison between \textit{coordinate-based} and \textit{voxel-based} representations.}
    \ra{1.2}
    \begin{tabular}{ccc}\toprule
    Representation & Runtime & \# Parameters ($L=128$)\\
    \midrule
    FourierNet & 0:09h & $\mathbf{330{,}240}$\\
    Voxel-grid & \textbf{0:06h} & $2{,}097{,}152$\\
    \bottomrule
    \end{tabular}
    \label{table:repcomparison}
\end{table}

\section{Avoiding Spurious Planar Symmetries}\label{supp:symm}
\subsection{Handedness Ambiguity in cryo-EM}
\label{suppsec:hand}

In cryo-EM, the interaction of the electron beam with the electrostatic potential $\vol$ in the orientation $\rot$ corresponds to an orthographic transparent projection described by
\begin{equation}
    Q_i = Q(\rot):(x,y) \mapsto \int_z \vol(\rot[x,y,z]^T)dz.
\end{equation}
For a set of projections $\{Q_i\}$, if the associated orientations are not given, there exists an instrinsic ambiguity on the handedness of the volume $\vol$~\cite{rosenthal2003optimal}. That is to say, one cannot distinguish a set of projections obtained with $\vol$ from any other set of projections obtained with a ``mirrored'' version of $\vol$.

More specifically, let us fix an orthonormal basis $\{\mathbf{e}_x, \mathbf{e}_y, \mathbf{e}_z\}$ on $\mathbb{R}^3$. We consider a volume $\vol:\mathbb{R}^3\to\mathbb{R}$ and define
\begin{equation}
    \tilde{\vol}(x,y,z) =  \vol(F[x,y,z]^T),
\end{equation}
where
\begin{equation}
    F = \begin{bmatrix}
    \phantom{a}1\phantom{a} & 0 & 0 \\
    0 & \phantom{a}1\phantom{a} & \phantom{a}0\phantom{a} \\
    0 & 0 & -1
    \end{bmatrix}.
\end{equation}
$\tilde{\vol}$ is the mirrored version of $\vol$ with respect to the $(x,y)$ plane. We also define the orientations $\rot$ using Euler angles in the ``$ZYZ$'' proper Euler convention. That is, each rotation matrix is parameterized by $\alpha_i$, $\beta_i$, $\gamma_i$ and
\begin{equation}
    \rot = R_{\alpha_i, \beta_i, \gamma_i} = \begin{bmatrix}
 c_1 c_3 + s_1 s_2 s_3 & \phantom{ab} c_3 s_1 s_2 - c_1 s_3 \phantom{ab} &   c_2 s_1 \\
  c_2 s_3 & c_2 c_3 & - s_2 \\
 c_1 s_2 s_3 - c_3 s_1 & c_1 c_3 s_2 + s_1 s_3 & c_1 c_2 
\end{bmatrix},
\end{equation}
where $c$ and $s$ represent cosines and sines (\textit{e.g} $c_1 = \cos \alpha_i$). Using this definition, we can show that
\begin{equation}
    F\rot = \tilde{\rot}F \quad \text{where } \tilde{\rot}=R_{\alpha_i+\pi, \beta_i, \gamma_i+\pi}.
\end{equation}
Therefore,
\begin{equation}
    \begin{split}
    \tilde{Q}_i = Q(\tilde{\rot}):(x,y)\mapsto & \int_z \tilde{\vol}(\tilde{\rot}[x,y,z]^T)dz\\
    & = \int_z \tilde{\vol}(\tilde{\rot}[x,y,-z]^T)dz \\
    & = \int_z \tilde{\vol}(\tilde{\rot}F[x,y,z]^T)dz \\
    & = \int_z \tilde{\vol}(F\rot[x,y,z]^T)dz \\
    & = \int_z \vol(\rot[x,y,z]^T)dz \\
    & = Q_i(x,y)
    \end{split}
\end{equation}
The second equality is a change of variable $z\to-z$. In conclusion, the volume $\vol$ associated with the set of orientations $\rot$ will give the same set of projections as the volume $\tilde{\vol}$ with the set of orientations $\tilde{\rot}$. For a symmetrical volume ($\vol=\tilde{\vol}$), $\rot$ and $\tilde{\rot}$ will give the same projections (Fig.~\ref{fig:halfsphere}, right).

\subsection{Spurious Planar Symmetries and Symmetric Loss}

\begin{figure}[h]
\centering
\includegraphics[width=0.9\textwidth]{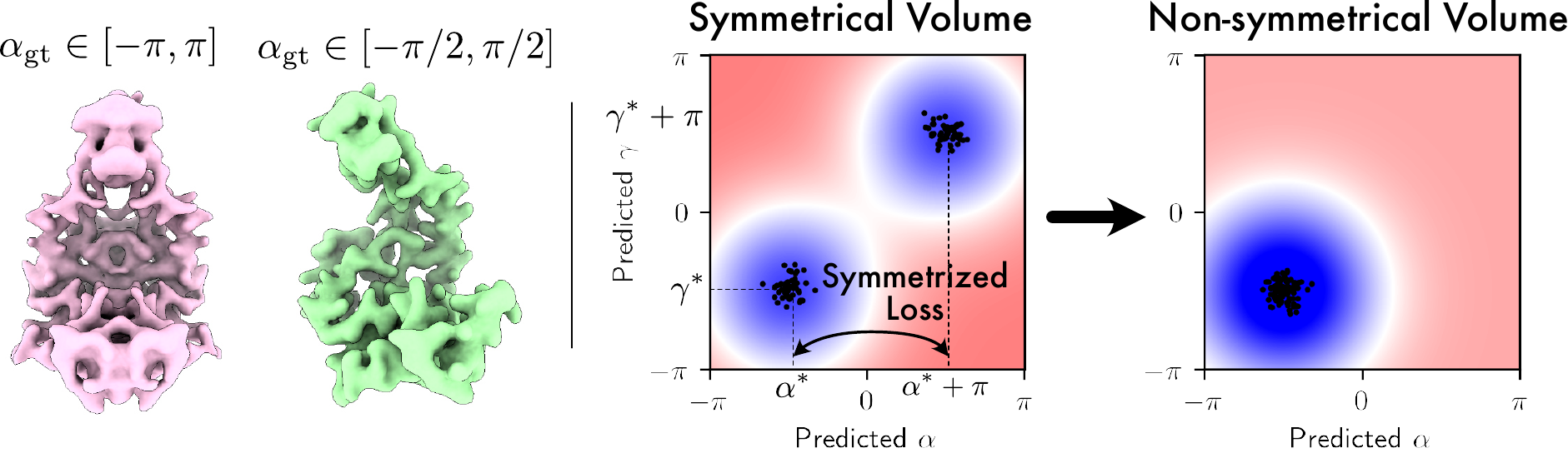}
\caption{(Left) Reconstructions obtained from the noisy adenylate kinase dataset ($\sidelen=128$) obtained with cryoAI and the L2 loss, depending on the range of simulated in-plane angles. The model gets stuck in a spuriously symmetrical state when poses are generated on all $SO(3)$ but does not when in-plane rotations are restricted to $[-\pi/2,\pi/2]$. (Right) Heatmap of the L2 loss (per image) and intuition on the role of the symmetric loss. With a symmetrical volume ($\vol=\tilde{\vol}$), the energy landscape is periodic and shows an energy barrier between two minima. Each black point qualitatively represents the predicted Euler angles for one image. Without symmetric loss, the model is stuck when the predicted angles are evenly distributed between the two minima. The symmetric loss helps the model to overcome the barrier by creating a ``tunnel'' in the energy landscape.}
\label{fig:halfsphere}
\end{figure}

\begin{figure}[t]
\centering
\includegraphics[width=\textwidth]{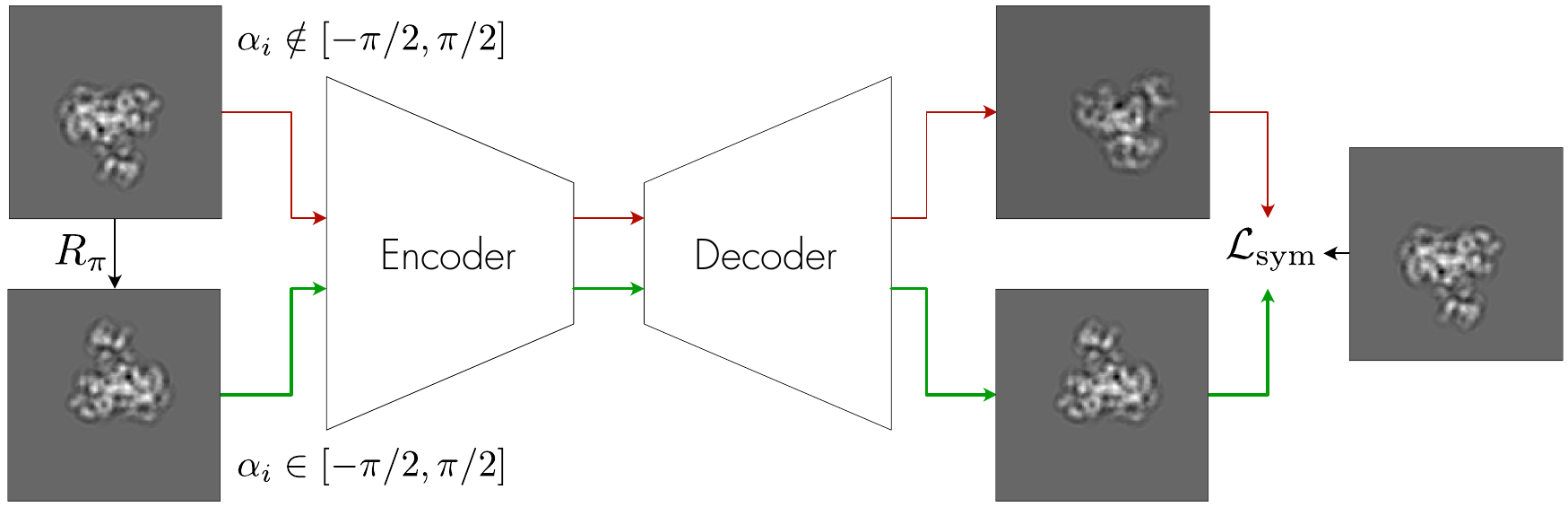}
\caption{For each input image, \cryoSIMBA produces two copies and rotates one them by $\pi$~rad. Two poses are predicted by the decoder and two images are reconstructed by the decoder. The symmetric loss only penalizes (in Fourier space) the lowest distance between the reconstructed images and the input image. During the backward pass, gradients are only backpropagated through the pass that gave the most accurate reconstruction (the green path here). Images are shown in real space for clarity.}
\label{fig:symmloss}
\end{figure}

When using our framework with a simple L2 loss
\begin{equation}
    \loss = \sum_{i\in\batch} \lVert \fprojgt - \fprojpred \rVert^2,
\end{equation}
we observed that the model got stuck in local minima where the volume was showing spurious symmetry planes (see Fig.~5). In order to verify that this behavior was linked to the ambiguity described in the previous section, we generated a simulated dataset with a $\alpha_i\in[-\pi/2, \pi/2]$ instead of $[-\pi, \pi]$. Doing so, the encoder only has to predict $\alpha_i$ in $[-\pi/2, \pi/2]$ (up to a rotation that does not depend on $i$) to get a correct reconstruction of the volume and will therefore be less likely to mis-classify the orientation $(\alpha_i, \beta_i, \gamma_i)$ as $(\alpha_i+\pi, \beta_i, \gamma_i+\pi)$. As shown in Fig.~\ref{fig:halfsphere}, our framework is able to accurately reconstruct $V$ with a L2 loss in that case.

We devised the symmetric loss (Eq.~(10)) using the observation that the model was able to quickly converge when supervised on images $\projgt$ for which $\alpha_i\in[-\pi/2, \pi/2]$.  As described in Fig.~\ref{fig:symmloss}, each image is firstly duplicated and one of the duplicates is rotated by $\pi$ (rotation layer in Fig.~\ref{fig:encoderarchi}). We know that one of the two duplicates has an associated in-plane angle $\alpha_i$ belonging to $[-\pi/2,\pi/2]$ (for some Euler basis). Each duplicate then goes through the whole pipeline and the two predicted images are compared with a L2 loss. The symmetric loss finds the minimum between the two L2 distances. This operation has the effect of disconnecting the loss from the worse predicted image in the computational graph. Therefore, at the backward pass, the gradients will only flow through one the of two paths, actually enabling the model to be supervised on images $\projgt$ for which $\alpha_i\in[-\pi/2, \pi/2]$, without any loss of information. Keeping track of which path was selected by the loss, we know which latent vector (rotation and translation) we should consider at the output of the encoder, when estimating the  poses.

Fig.~\ref{fig:halfsphere} (right) gives an intuition on the role of the symmetric loss. For a symmetrical volume ($\vol=\tilde{\vol}$), the L2 loss is  periodic for each image ($\loss(\alpha,\gamma)=\loss(\alpha+\pi,\gamma+\pi)$, see~\ref{suppsec:hand}). Therefore, there exists two identical global minima in the energy landscape. The model is stuck when predicted angles are distributed evenly between the two minima. The symmetric loss creates a ``tunnel'' in the energy landscape, helping the model to overcome the energy barrier and distribute the predicted angles in a way that is consistent with a non-symmetrical volume.

\subsection{Symmetric Loss on PoseVAE}

\begin{figure}[h]
\centering
\includegraphics[width=0.5\textwidth]{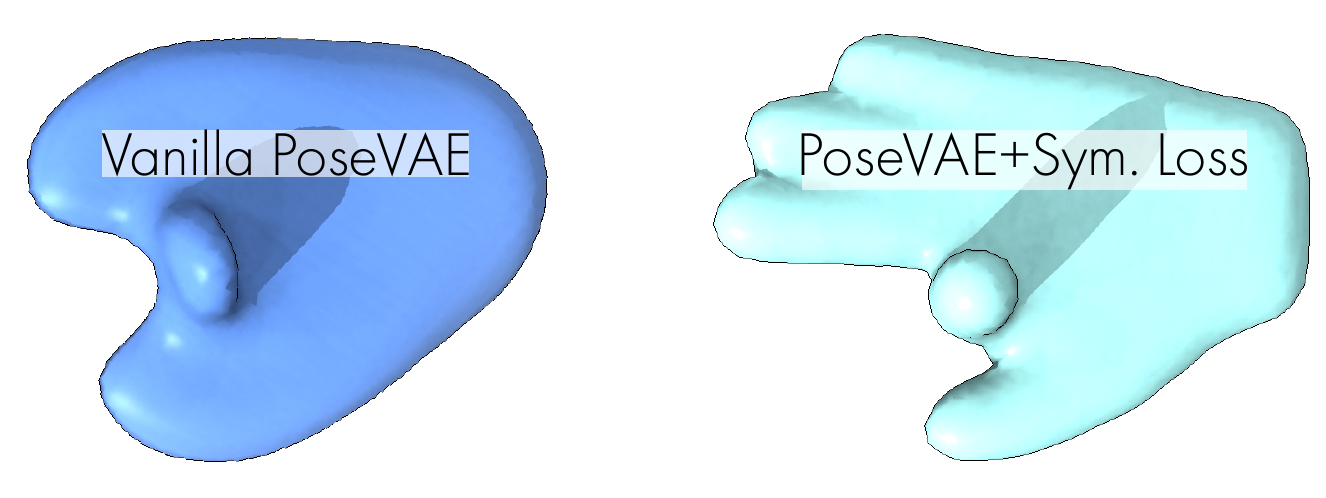}
\caption{Comparison between L2 loss (vanilla) and the symmetric loss with the PoseVAE method on the \textit{hand pointer} dataset, both from \cite{zhong_cryodrgn2_2021}.}
\label{fig:hand}
\end{figure}

In \cite{zhong_cryodrgn2_2021}, Zhong \textit{et. al.} proposed to use a variational auto-encoder to predict the pose and named their technique ``PoseVAE''. Visual reconstruction show that their reconstructed volume gets stuck in symmetric states. We reproduced their method and added the symmetric loss thereby enabling PoseVAE to work on the \textit{hand pointer} dataset (Fig.~\ref{fig:hand}).

\section{Datasets}\label{supp:datasets}
\begin{figure}[h!]
\centering
\includegraphics[width=0.85\textwidth]{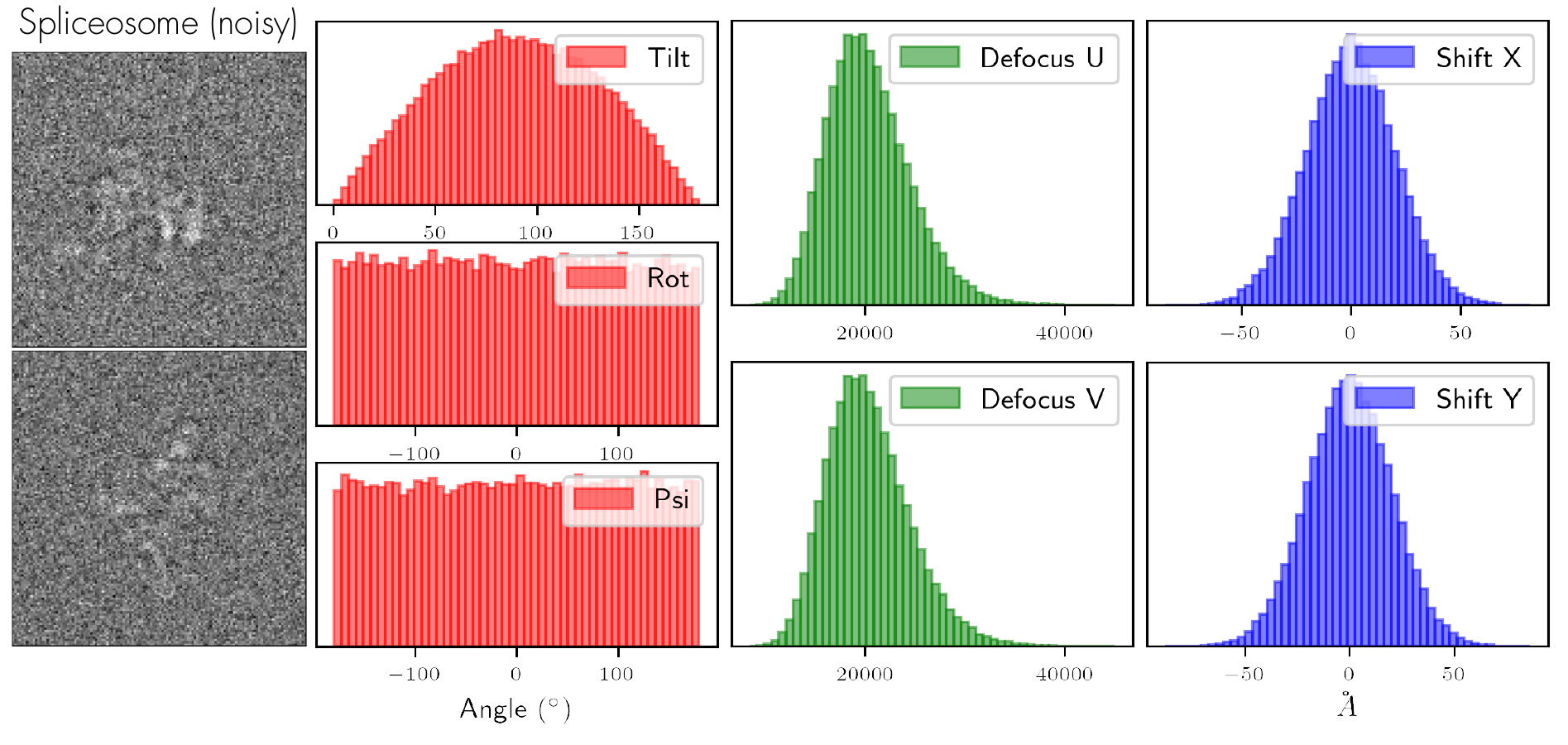}
\caption{Statistics and samples from the simulated noisy spliceosome dataset.}
\label{fig:spliceosomestatsshort}
\end{figure}

\begin{figure}[h!]
\centering
\includegraphics[width=0.85\textwidth]{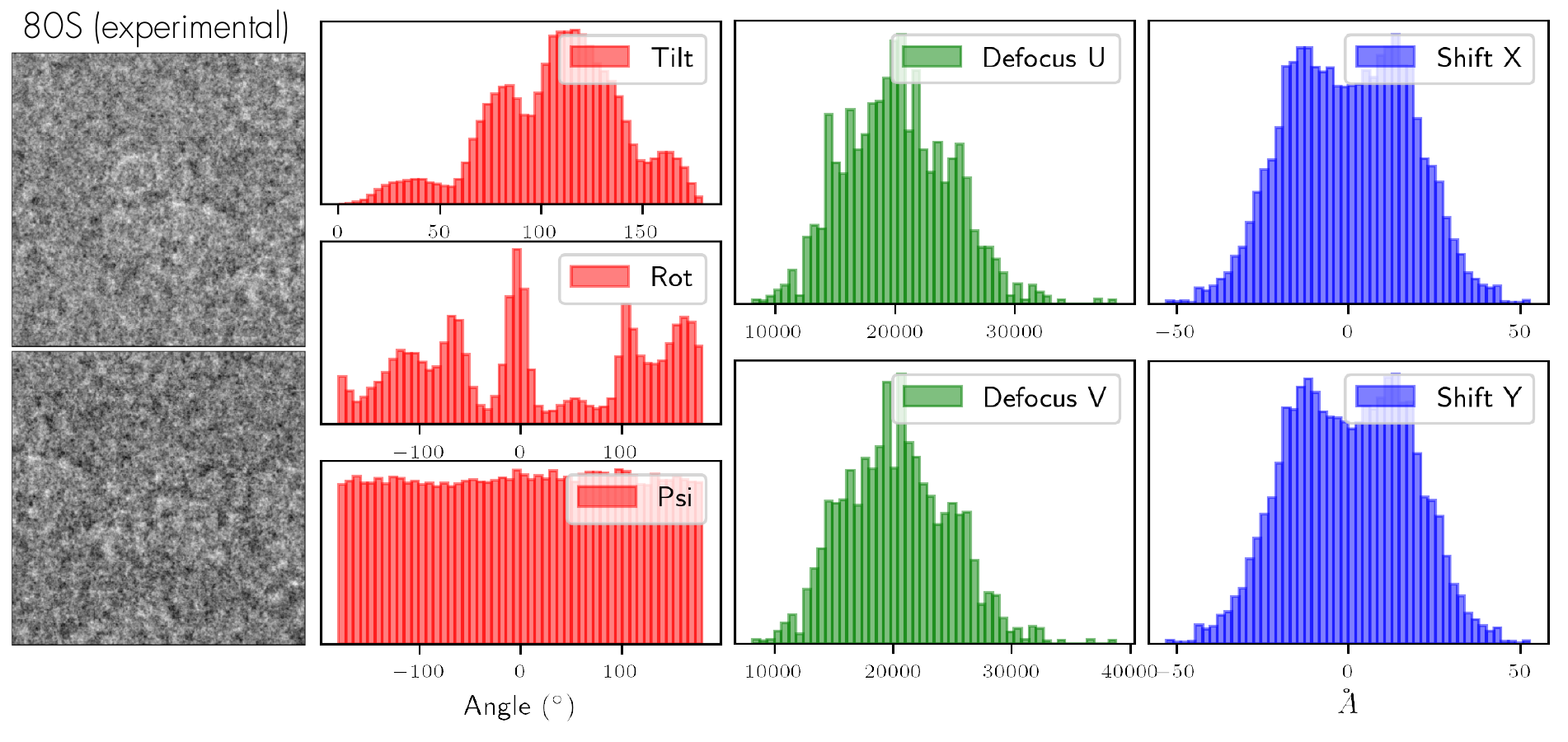}
\caption{Statistics and samples from the experimental 80S dataset.}
\label{fig:80Sstats}
\end{figure}

Table~\ref{table:datasetstats} summarizes the parameters of the simulated and experimental datasets we used. Fig.~\ref{fig:spliceosomestatsshort} (resp. Fig.~\ref{fig:80Sstats}) shows in more details the statistics of the poses, the defoci and the shifts in the dataset for the simulated spliceosome (resp. experimental 80S).

\begin{table}[h!]
    \scriptsize
    \centering
    \caption{Parameters for our datasets. $\sidelen$ is the image size, $N$ is the number of images in the dataset. We give samples of the datasets in Fig.~\ref{fig:spliceosomestatsshort}~and~\ref{fig:80Sstats} to show the level of noise.}
    \ra{1.2}
    \begin{tabular}{llcccccccccccc}\toprule
    \multicolumn{2}{c}{Dataset} & $\sidelen$ & \phantom{abc} & $N$ & \phantom{abc} & \r{A}/pix. & \phantom{abc} & Shift? & \phantom{abc} & SNR (dB)\\
    \midrule
    \textit{Simulated} & {80S noisy} & $128$ && $10\text{k-}9\text{M}$ && $3.77$ && N && $0$\\
    & {Spike ideal} & $128$ && $50{\small,}000$ && $3.00$ && N && $\infty$\\
    & {Spike noisy} & $128$ && $50{\small,}000$ && $3.00$ && N && $-10$\\
    & {Spliceosome ideal} & $128$ && $50{\small,}000$ && $4.25$ && Y && $\infty$\\
    & {Spliceosome noisy} & $128$ && $50{\small,}000$ && $4.25$ && Y && $-10$\\
    & {Kinase ideal} & $64$ && $10{\small,}000$ && $3.20$ && N && $\infty$\\
    & {Kinase noisy} & $128$ && $10{\small,}000$ && $3.20$ && N && $-5$\\
    \textit{Experimental} & {80S EMPIAR-10028} & $256$ && $105{\small,}247$ && $1.89$ && Y && NA\\
    \bottomrule
    \end{tabular}
    \label{table:datasetstats}
\end{table}

\section{Additional Results}\label{supp:addres}
\subsection{Full Evaluation of Poses}

\begin{figure}[t]
    \centering
    \includegraphics[width=0.9\textwidth]{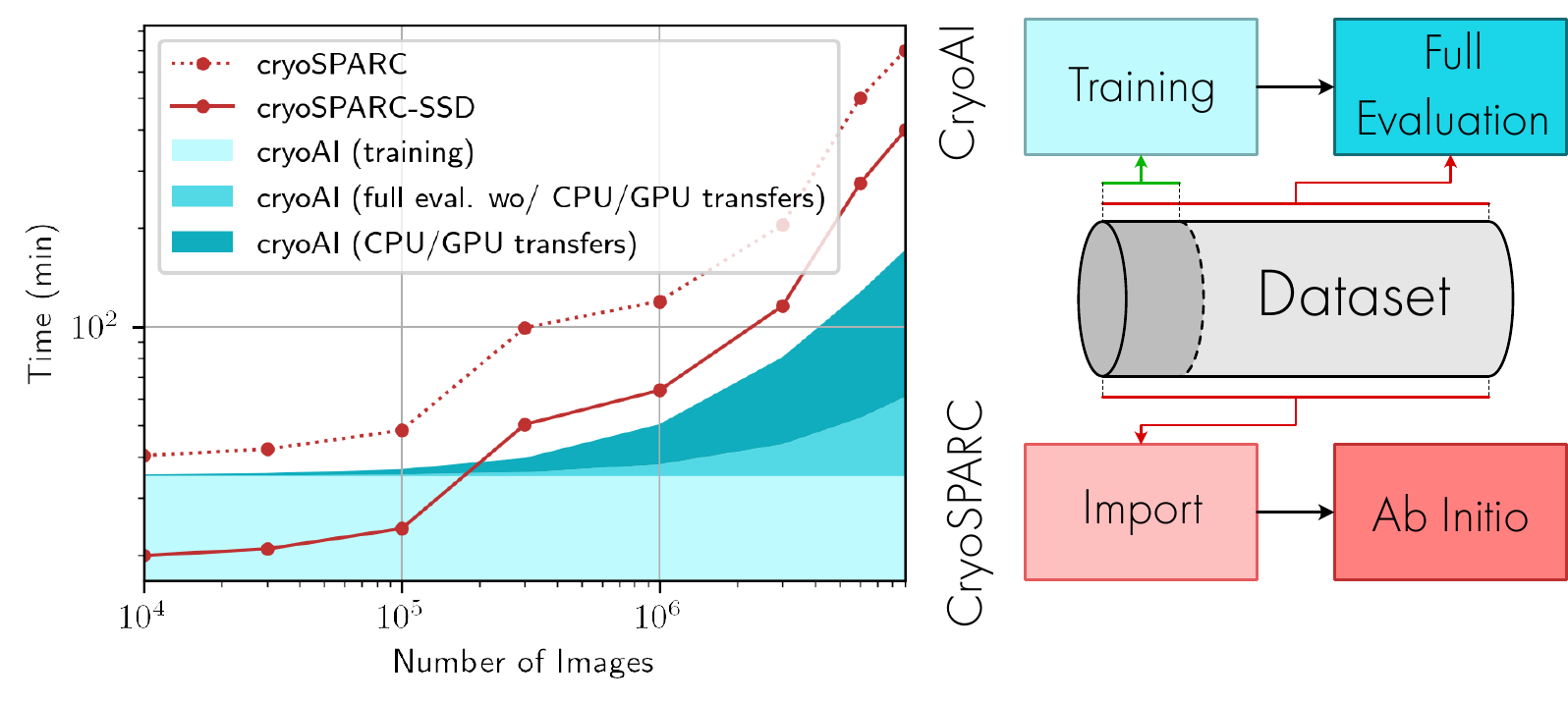}
    \caption{(Left) Time required to estimate all the poses in the simulated noisy 80S dataset, with cryoAI and cryoSPARC. CryoAI switches to evaluation mode once a resolution of $10$~\r{A} is achieved on the volume. We indicate the time spent transferring data between the CPUs and the GPU during evaluation with cryoAI. CryoSPARC-SSD speeds up computations by caching particles on local SSDs. (Right) Comparison of workflows. CryoAI converges before seeing the whole dataset and can process rapidly the remaining images in evaluation mode. CryoSPARC imports (and formats) the dataset before processing it.}
    \label{fig:timefulleval}
\end{figure}

We compared in Fig.~3 the runtime of cryoAI and cryoSPARC to reach a resolution of $10$~\r{A} on the reconstructed volume with the simulated noisy 80S dataset. In that experiment, cryoAI may reach convergence before processing all images in the dataset. For the sake of completeness, we run a second experiment, whose results are shown in Fig.~\ref{fig:timefulleval}. Once cryoAI has reached the threshold resolution of $10$~\r{A}, we use our model in evaluation mode (the computational graph is not maintained anymore, which decreases the memory cost of a forward pass and enables us to increase the batch size) to predict the poses associated with all images in the dataset. We report the time required to evaluate the whole dataset with cryoAI and with cryoSPARC. With cryoAI, we show the time required for CPU-to-GPU and GPU-to-CPU data transfers since this time could potentially be compressed with smart data handling. When a solid-state drive (SSD) is available, cryoSPARC can significantly decrease the time of \emph{ab initio} reconstruction with particle caching. We can deduce from Fig.~S8, that the runtime per image is $1.1$~ms (2:35h / 9M) for cryoAI vs. $4.7$~ms (11:36h / 9M) for cryoSPARC ab initio without SSD, which goes to show the computational benefits of using an encoder to estimate poses.

\subsection{Experimental 80S}

\begin{figure}
    \centering
    \includegraphics[width=\textwidth]{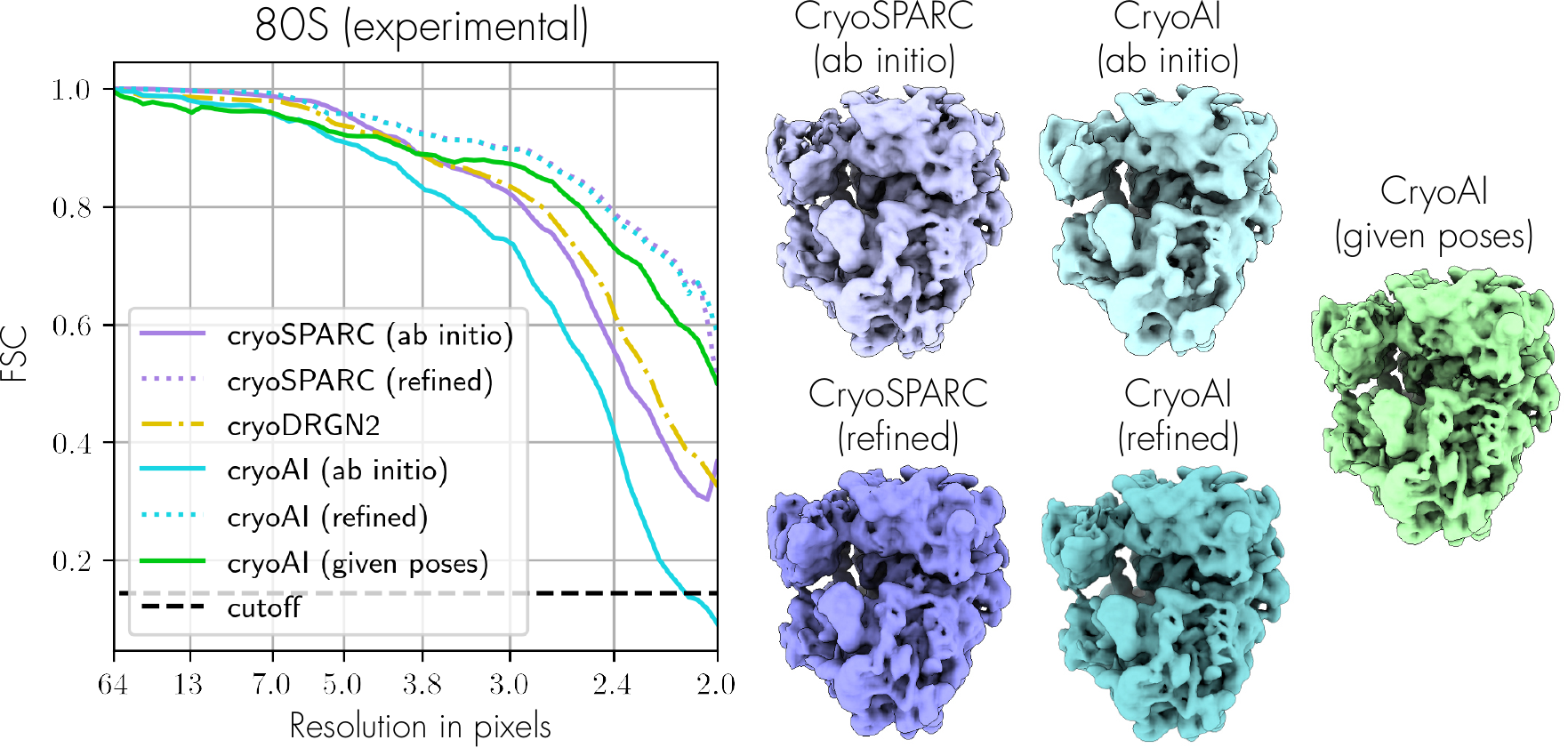}
    \caption{(Left) FSC reconstruction-to-reconstruction on the experimental dataset of 80S. (Right) Reconstructed volumes visualized with UCSF ChimeraX~\cite{pettersen2021ucsf}.}
    \label{fig:80Ssupp}
\end{figure}

\textbf{Additional Results.} Table~\ref{table:expres} and Fig.~\ref{fig:80Ssupp} show quantitative and qualitative results obtained with the experimental dataset of the 80S. In the absence of a ground truth volume, we perform a volume reconstruction with cryoAI using published poses.

\textbf{Input resolution.} We observed that cryoAI did not properly converge when fed with input images of size $\sidelen=128$. Increasing the input size to $256$ provides more information to the encoder for pose estimation but also implies making $256^2$ queries per image to the FourierNet. For the computation to fit on a $40$~Gb GPU, we need to decrease the batch size to $8$, which makes the gradient-descent too stochastic and prevents the model from converging. Our solution was to use an image size of $256$ on the encoder size and keep an image size of $128$ on the decoder size. Once convergence is reached, voxel grids of sizes $256^3$ or $128^3$ can be queried in the FourierNet. However, in order to stay consistent with the image size used during training, we chose to output volumes of sizes $128^3$. For the comparison with cryoSPARC to be fair, we use images of sizes $256$ and downsample the volumes reconstructed by cryoSPARC from $256^3$ to $128^3$. CryoDRGN2 only reports the results obtained with $\sidelen=128$.

\begin{table}[t]
    \scriptsize
    \centering
    \caption{Accuracy of pose and volume estimation for experimental 80S data. Resolution (Res.) is reported using the $\text{FSC}=0.143$ criterion, in \r{A} $(\downarrow)$. Rotation (Rot.) error is the median square Frobenius norm between predicted and published poses matrices $\rot$ $(\downarrow)$. Translation (Trans.) error is the mean square L2-norm, in \r{A} $(\downarrow)$. ``\cryoSIMBA + cryoSPARC'' refers to \cryoSIMBA (\emph{ab initio}) + cryoSPARC (refinement).}
    \ra{1.2}
    \begin{tabular}{rcccccccccc}\toprule
    \textit{80S (exp.)} & \phantom{abc} & cryoSPARC & \phantom{abc} & cryoDRGN2 & \phantom{abc} & \cryoSIMBA & \phantom{abc} & \cryoSIMBA + cryoSPARC\\
    \midrule
    Res. \r{A} && $\mathbf{7.54}$ && $\mathbf{7.54}$ && $7.91$ && $\mathbf{7.54}$\\
    Rot. && $\mathbf{0.0001}$ && $0.0008$ && $0.004$ && $\mathbf{0.0001}$\\
    Trans. && $\mathbf{0.0008}$ && $0.002$ && $0.005$ && $\mathbf{0.0008}$\\
    \bottomrule
    \end{tabular}
    \label{table:expres}
\end{table}

\subsection{Experimental Precatalytic Spliceosome}

\begin{figure}
    \centering
    \includegraphics[width=\textwidth]{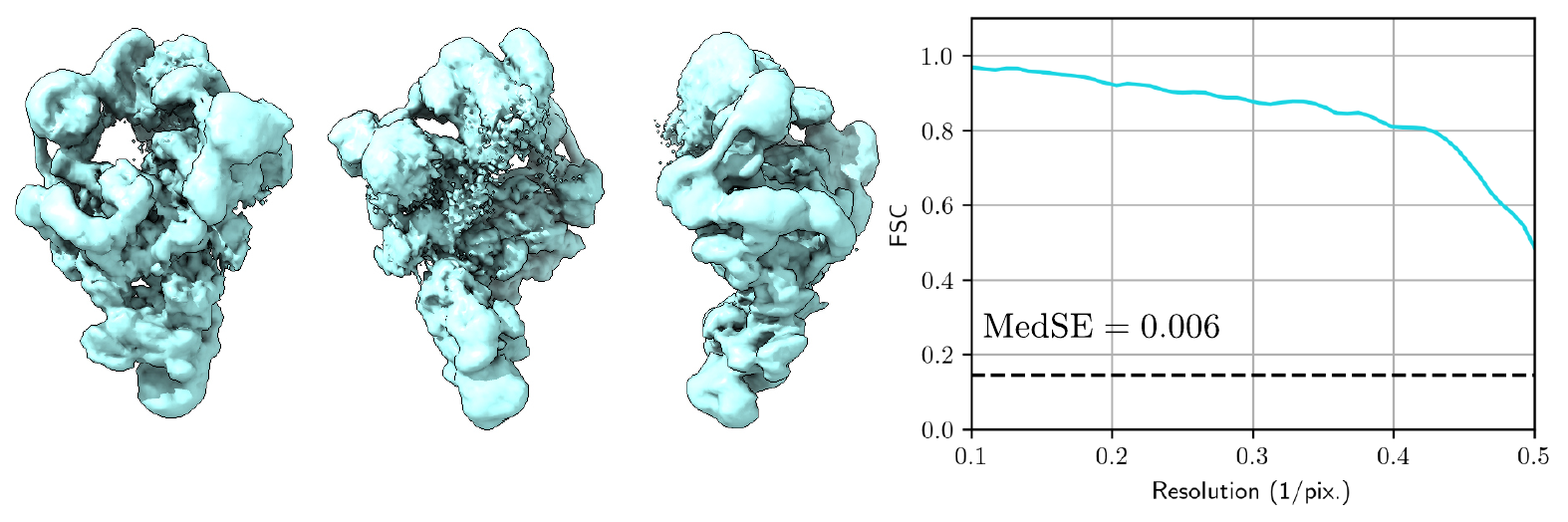}
    \caption{Qualitative reconstruction on the EMPIAR-10180 dataset~\cite{cryodrgn_zenodo}.}
    \label{fig:empiar10180}
\end{figure}

We downloaded images of the precatalytic spliceosome from EMPIAR-10180~\cite{spliceosome}, and downsampled to $D=128$ (4.25 \AA/pix). We performed homogeneous reconstruction with cryoAI on the filtered set of 139,722 images available at~\cite{cryodrgn_zenodo}. Particle images are shifted by their published poses, since the particles in this dataset are significantly out of center~\cite{zhong_cryodrgn2_2021}. Results are shown in Fig.~\ref{fig:empiar10180}. We report the half-to-half FSC. We note the presence of a blurry zone in the reconstructed volume, which correlates with the zone where the molecule can fold.

\subsection{Simulated Datasets}

\begin{figure}[h!]
    \centering
    \includegraphics[width=\textwidth]{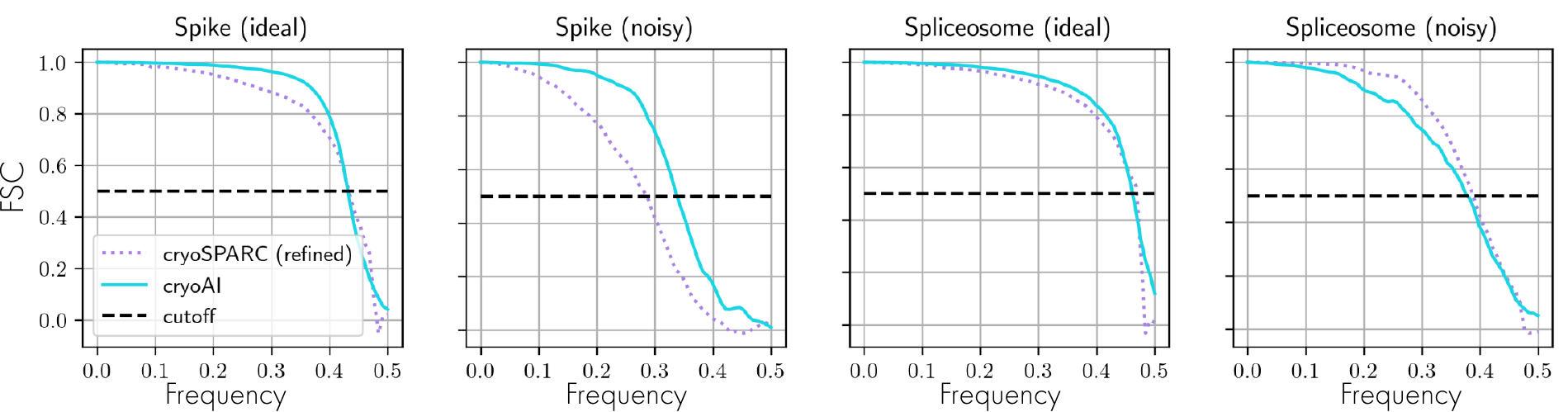}
    \caption{Fourier Shell Correlations reconstruction-to-ground-truth on simulated datasets, with a cutoff at $\text{FSC}=0.5$.}
    \label{fig:simsupp}
\end{figure}

\begin{figure}[h!]
    \centering
    \includegraphics[width=0.63\textwidth]{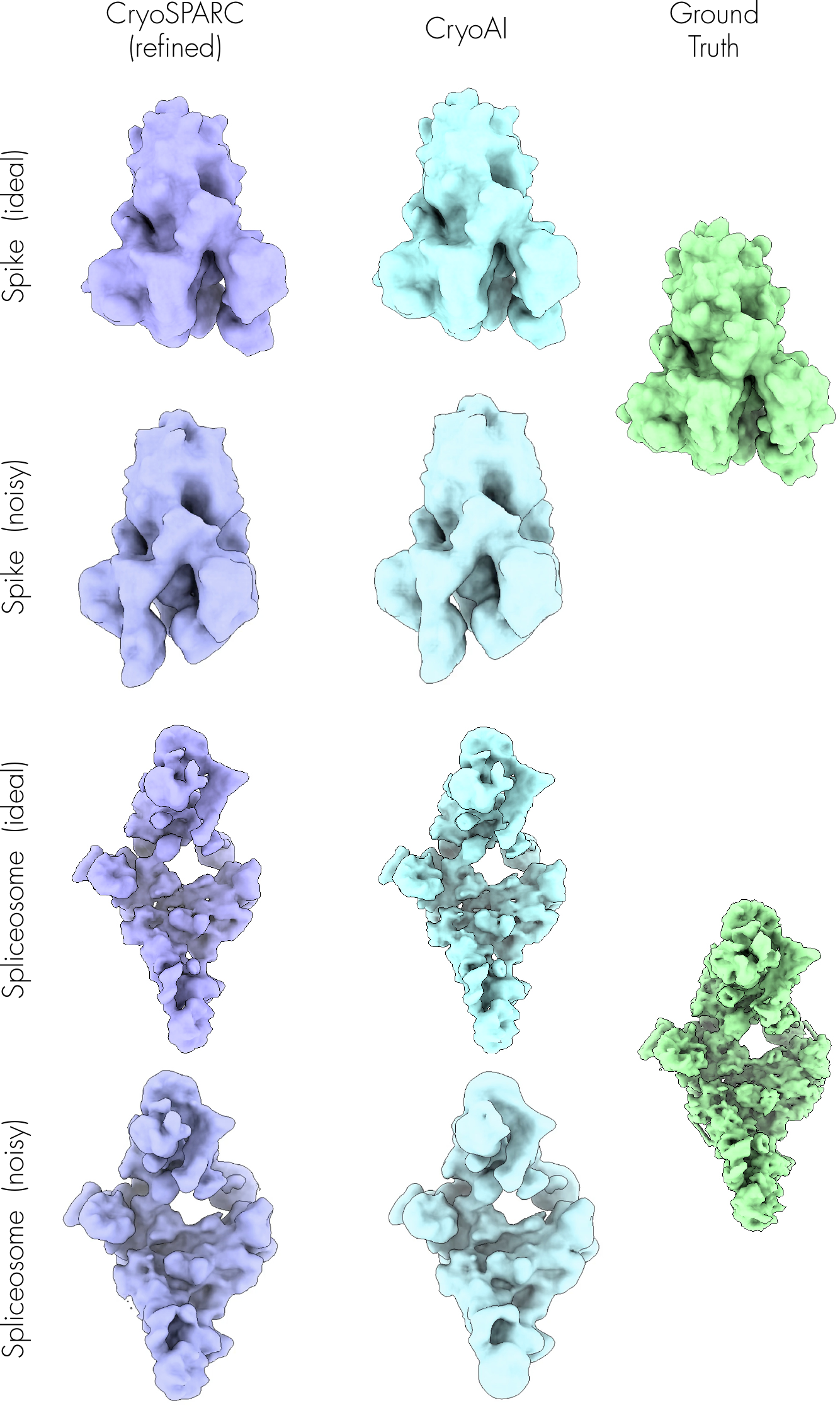}
    \caption{Qualitative reconstructions on simulated datasets.}
    \label{fig:qualisimsupp}
\end{figure}

We show in Fig.~\ref{fig:simsupp}-\ref{fig:qualisimsupp} quantitative and qualitative results obtained with our simulated datasets with cryoAI and cryoSPARC. We quantitatively study the impact on the noise level on a small (10k images) synthetic dataset of the 80S ribosome ($L=128$) in Table~\ref{table:impactnoise}. We can see that the convergence time increases with the noise while the final resolution decreases slightly.

\begin{table}
    \centering
    \caption{Impact of the noise on the runtime and the resolution with the synthetic 80S dataset.}
    \ra{1.2}
    \begin{tabular}{ccccc}\toprule
    SNR & $0$dB & $-5$dB & $-10$dB & $-15$dB\\
    \midrule
    Runtime & 0:20h & 0:29h & 0:45h & 2:07h \\
    Res. (pix) & 2.15 & 2.44 & 2.79 & 3.21\\
    \bottomrule
    \end{tabular}
    \label{table:impactnoise}
\end{table}

Molecular graphics and analyses performed with UCSF ChimeraX~\cite{pettersen2021ucsf}, developed by the Resource for Biocomputing, Visualization, and Informatics at the University of California, San Francisco, with support from National Institutes of Health R01-GM129325 and the Office of Cyber Infrastructure and Computational Biology, National Institute of Allergy and Infectious Diseases.

\newpage

\bibliographystyle{splncs04}
\bibliography{supplementary}